\documentclass{article}

\PassOptionsToPackage{numbers, compress}{natbib}

    \usepackage[preprint]{neurips_2025}

\usepackage{amsmath,amsfonts,bm}

\def\eqref#1{equation~\ref{#1}}

\def\1{\bm{1}}

\DeclareMathAlphabet{\mathsfit}{\encodingdefault}{\sfdefault}{m}{sl}
\SetMathAlphabet{\mathsfit}{bold}{\encodingdefault}{\sfdefault}{bx}{n}

\newcommand{\bc}{\mathbf{c}}

\newcommand{\bv}{\mathbf{v}}

\newcommand{\bx}{\mathbf{x}}

\newcommand{\bz}{\mathbf{z}}

\newcommand{\defeq}{\vcentcolon=}

\usepackage[utf8]{inputenc} 
\usepackage[T1]{fontenc}    
\usepackage{hyperref}       
\usepackage{url}            
\usepackage{booktabs}       
\usepackage{amsfonts}       
\usepackage{nicefrac}       
\usepackage{microtype}      
\usepackage{xcolor}         

\usepackage{algorithm}
\usepackage{algorithmic}
\usepackage{amsmath}
\usepackage{mathtools}
\usepackage{multirow}
\usepackage{color, colortbl}

\usepackage{caption}

\usepackage[accsupp]{axessibility} 
\usepackage{times}
\usepackage{epsfig}
\usepackage{amssymb}
\usepackage{etoc}
\usepackage{appendix}
\usepackage{minitoc}
\usepackage[para]{footmisc}
\usepackage{caption}
\usepackage{natbib}
\usepackage{wrapfig,lipsum,booktabs}
\usepackage{titletoc}
\newcommand{\methodname}{{ARFlow}}

\title{\methodname: Human Action-Reaction Flow Matching with Physical Guidance}

\newcommand*{\affmark}[1][*]{\textsuperscript{#1}}

\author{%
  Wentao Jiang$^{1}$ \quad
  Jingya Wang$^{1*}$ \quad
  Kaiyang Ji$^{1}$ \quad
  Baoxiong Jia$^{2}$ \quad
  Siyuan Huang$^{2}$ \quad
  Ye Shi$^{1}\thanks{Corresponding authors.}$\\
  \\
  \affmark[1]{ShanghaiTech University} \quad
  \affmark[2]{Beijing Institute for General Artificial Intelligence} \\
   \\
  \texttt{\{jiangwt2024,wangjingya,jiky2024,shiye\}@shanghaitech.edu.cn} \\
  \texttt{\{jiabaoxiong,syhuang\}@bigai.ai} \\  
}

\begin{document}

\maketitle


\begin{abstract}
Human action-reaction synthesis, a fundamental challenge in modeling causal human interactions, plays a critical role in applications ranging from virtual reality to social robotics. While diffusion-based models have demonstrated promising performance, they exhibit two key limitations for interaction synthesis: reliance on complex noise-to-reaction generators with intricate conditional mechanisms, and frequent physical violations in generated motions. To address these issues, we propose Action-Reaction Flow Matching (ARFlow), a novel framework that establishes direct action-to-reaction mappings, eliminating the need for complex conditional mechanisms. Our approach introduces a physical guidance mechanism specifically designed for Flow Matching (FM) that effectively prevents body penetration artifacts during sampling. Moreover, we discover the bias of traditional flow matching sampling algorithm and employ a reprojection method to revise the sampling direction of FM. To further enhance the reaction diversity, we incorporate randomness into the sampling process. Extensive experiments on NTU120, Chi3D and InterHuman datasets demonstrate that ARFlow not only outperforms existing methods in terms of Fréchet Inception Distance and motion diversity but also significantly reduces body collisions, as measured by our new Intersection Volume and Intersection Frequency metrics. Our project is available at \url{ https://arflow2025.github.io/}. 
\end{abstract}

\section{Introduction}
\label{sec:intro}
Human action-reaction synthesis~\cite{tanthink,xu2024regennet,chopin2023interaction} has emerged as a pivotal research direction in computer vision~\cite{actformer, intergen, starke2020local, javed2024intermask, role_aware, wang2023intercontrol}. This task aims to generate physically plausible human reactions responding to observed actions, with critical applications in virtual reality, human-robot interaction, and character animation. Unlike single-human motion generation~\cite{action2motion,tevet2022human_mdm,chen2023executing_mld}, reactors must infer responses without observing future actor motions, creating unique modeling challenges. 

While recent diffusion methods~\citep{tevet2022human_mdm,xu2024regennet} show promise in motion generation, they face two key limitations in the modeling of action-reaction interactions. First, existing approaches ~\citep{tevet2022human_mdm,xu2024regennet} indirectly model responses using noise-to-reaction generators with intricate conditional mechanisms like treating action information as a condition to guide the generation process. This not only complicates the training process but also fails to directly capture the inherent causal relationship between actions and reactions. Second, frequent physical violations like body penetration between characters occur due to neglected physical constraints. While such issues are absent in single-human scenarios, they become critical in human interaction applications~\citep{hoyet2012push}. This poses a significant barrier to real-world applications such as virtual reality and human-robot interaction, where even minor physical inaccuracies are intolerable~\cite{reitsma2003perceptual, hoyet2012push}. 

\begin{figure*}[t]
  \vspace{-2mm}
  \centering
  \includegraphics[width=0.9\linewidth]{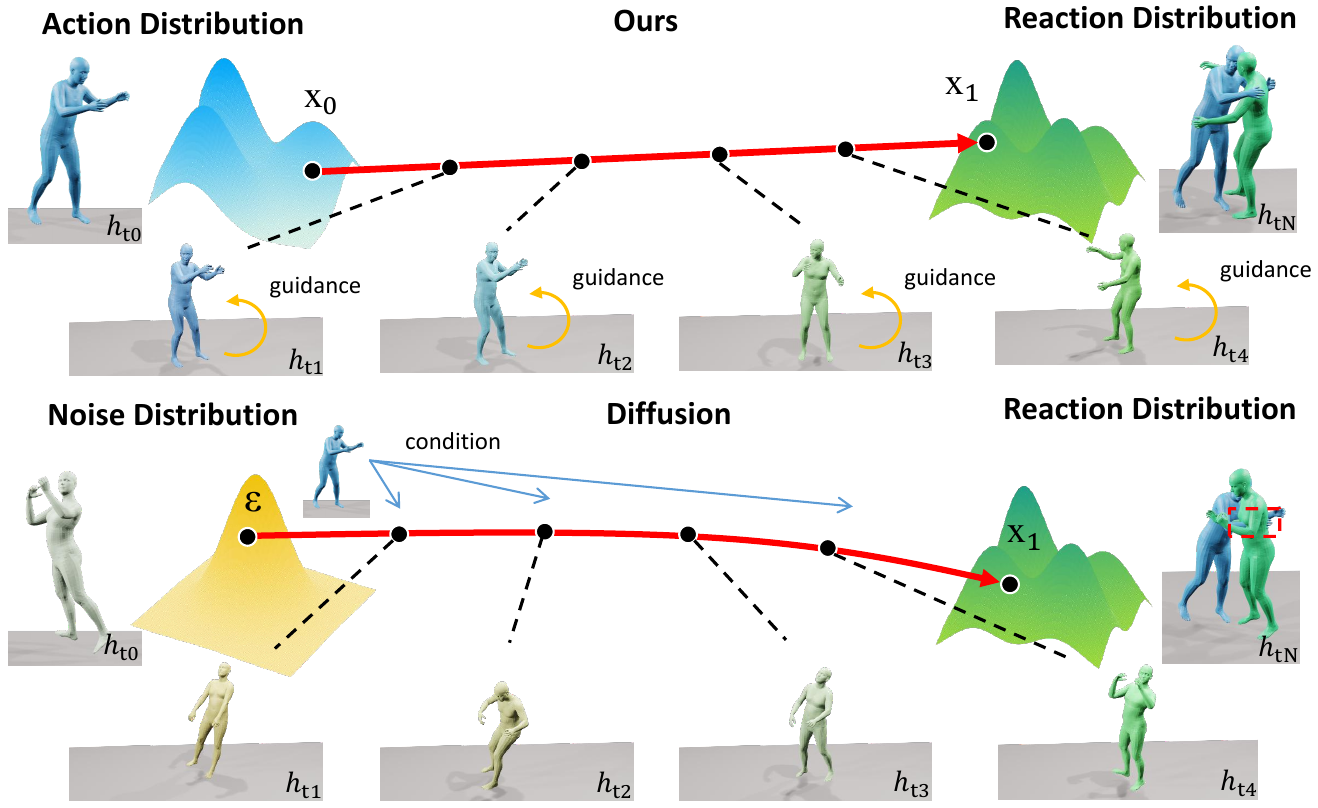}
  \vspace{-2mm}
  \caption{Our proposed Human Action-Reaction Flow (\textbf{\methodname}). We directly establish a mapping between the action and reaction distribution and our sampling process is further guided by our physical constraint guidance. The change of colors represents the variation of the $h$-frame reaction with sampling timestep $t_n$. }
  \label{fig:colour}
  \vspace{-2mm}
\end{figure*}

To address these challenges, we propose Action-Reaction Flow Matching (ARFlow), a novel framework that fundamentally resolves these limitations. Unlike diffusion models constrained by noise-data mappings, flow matching~\citep{lipman2022flow,rectifiedflow_iclr23,albergo2022building,neklyudov2023action} naturally models paired distributions through linear interpolation between endpoints~(See Fig.~\ref{fig:colour}), making it inherently suitable for action-reaction synthesis. ARFlow establishes direct pathways between action and reaction distributions, enabling simpler training and faster inference compared to diffusion-based methods. Instead of traditional vector field prediction ($\bv$-prediction) in standard flow matching~\citep{lipman2022flow, hu2023motion}, ARFlow directly outputs clean body poses ($\bx_1$-prediction) in human motion generation. To eliminate unrealistic body collisions between characters, we further develop a specialized sampling algorithm that performs physical constraint guidance at $\bx_1$ to avoid inaccurate gradients, and projects it back onto a revised Flow Matching path through interpolation. This innovation maintains physical plausibility through gradient guidance without compromising motion quality.
Additionally, we introduce two new physics-aware evaluation metrics: Intersection Volume (IV) and Intersection Frequency (IF). Our main contributions are threefold: 
\begin{itemize}
    \item We propose ARFlow, the first Flow Matching architecture that creates direct pathways between human action and reaction distributions, eliminating the need for complex conditional mechanisms prevalent in diffusion approaches. 
    \item We integrate physical constraints and employ a reprojection method to revise the sampling direction of FM, surpassing vanilla guidance methods. This successfully prevents body collisions between characters while preserving natural movement. To further enhance the reaction diversity, we introduce randomness into the sampling process.
    \item We introduce two new evaluation metrics (IV, IF) for interaction quality assessment. Experiments demonstrate state-of-the-art performance in Fréchet Inception Distance (FID), MultiModality (Multimod.), IV and IF on NTU120 and Chi3D datasets. 
\end{itemize}

\section{Related Work}
\label{sec:related}
\vspace{-5pt}
\subsection{Human Motion Generation}
\vspace{-5pt} 

Human motion synthesis aims to generate diverse and realistic human-like motion conditioned on different guidances, including text~\cite{zhang2023remodiffuse,guo2022tm2t,flame,mofusion}, music~\cite{zhou2023ude,li2022danceformer}, speech~\cite{habibie2022motion,ao2022rhythmic}, sparse signals \cite{agrol, tang2024unified}, action labels~\cite{action2motion,petrovich2021action,tevet2022human_mdm,cervantes2022implicit,chen2023executing_mld}, 
or unconditioned~\cite{modi,tevet2022human_mdm,csgn}. Recently, many diffusion-based motion generation models have been proposed~\cite{motiondiffuse,tevet2022human_mdm,mofusion,chen2023executing_mld,wu2024thor,tian2024gaze} and demonstrate better quality compared to alternative models such as
VAE~\cite{vae,action2motion,petrovich2021action,cervantes2022implicit}, flow-based models~\cite{rezende2015variational,aliakbarian2022flag} or GANs~\cite{csgn,actformer}.
Alternatively, motion can be regarded as a new form of language
and embedded into the language model framework~\cite{zhang2023generating_t2mgpt,jiang2023motiongpt}.

Meanwhile, the exploration of guiding the sampling process of diffusion models \cite{chung2023diffusion,yang2024guidance} has been a key area in motion diffusion models, PhysDiff~\cite{karunratanakul2023guided} proposes a physics-guided motion diffusion model, which incorporates physical constraints in a physics simulator into the diffusion process. GMD~\cite{karunratanakul2023guided} presents methods to enable spatial guidance without retraining the model for a new task. DNO~\cite{karunratanakul2023guided} proposes a motion editing and control approach by optimizing the diffusion latent noise of an existing pre-trained model.

\vspace{-5pt}
\subsection{Human Action-Reaction Synthesis}
\vspace{-5pt}
Different from human-human interaction~\cite{actformer, intergen, starke2020local, javed2024intermask, role_aware, wang2023intercontrol}, human action-reaction synthesis is causal and
asymmetric~\cite{xu2024regennet,nturgbd120, xu2023inter}. To address this task, researchers have leveraged large language models~\cite{siyao2024duolando, tanthink, jiang2024solami}, VAE-based methods~\cite{chopin2023interaction,liu2023interactive,liu2024physreaction}. 
However, these methods cannot capture fine-grained representations and ensure diversity,
and diffusion-based methods~\cite{li2024interdance,role_aware} are limited to the ``offline'' and ``constrained'' setting of human reaction generation, failing to generate instant and intention agnostic reactions. More recently, ReGenNet~\cite{xu2024regennet} introduce a diffusion-based transformer decoder framework and treat action sequence as conditional signal for online reaction generation. However, it often produce physically-implausible inter-penetrations between the actor and reactor since they disregard physical constraints in the generative process. Our method addresses this problem by ARFlow sampling with physical constraint guidance. 

\vspace{-5pt} 
\subsection{Flow Matching}
\vspace{-5pt} 
Flow Matching~\citep{lipman2022flow, rectifiedflow_iclr23,albergo2022building,neklyudov2023action,martin2024pnp,feng2025guidance} has emerged as an efficient alternative to diffusion models, offering linear generation trajectories through ODE solvers. This paradigm enables simplified training and accelerated inference~\citep{esser2024scaling,lipman2024flow}, with successful applications spanning images~\citep{lipman2022flow,esser2024scaling}, audio~\citep{le2023voicebox}, video~\citep{video_fm}, and point clouds~\citep{wu2023fast}. In motion generation, MotionFlow~\citep{hu2023motion} demonstrates comparable performance to diffusion models with faster sampling. Notably, Flow Matching inherently models transitions between arbitrary distributions through transport maps, making it particularly suitable for paired data modeling. Despite these advantages, its potential for action-reaction synthesis remains unexplored. Our work bridges this gap by establishing direct action-to-reaction mappings without complex conditional mechanisms.

\section{Method}
\label{sec:flow}
\begin{figure*}[t]
  \vspace{-2mm}
  \centering
  \includegraphics[width=1.0\linewidth]{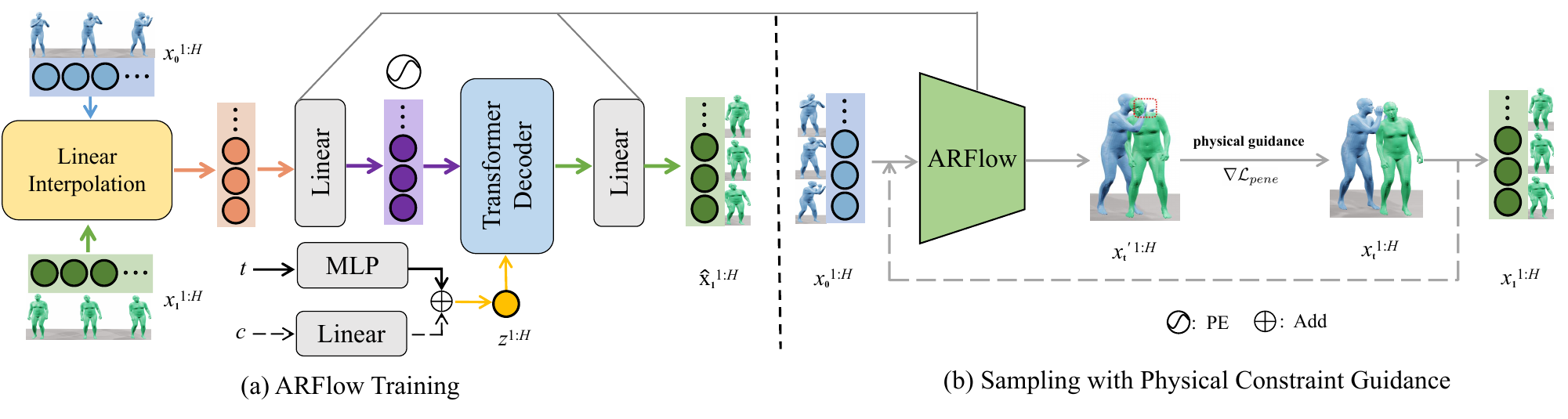}
  \vspace{-2mm}
  \caption{\textbf{Pipeline of \methodname.} \textbf{(a)} For a sampled timestep $t$, we linearly interpolate a coupled action-reaction pair as Eq.~\ref{eqn:flow} to produce the intermediate state $x^{1:H}_t$, which is then turns into a \textit{d}-dimensional latent feature through a linear layer. We use Transformer Decoder Units to directly predict clean reaction motions. \textbf{(b)} After training the networks in (a), our ARFlow uses them for sampling, which is further guided by the gradient of $\mathcal{L}_{pene}$ to generate physically plausible reactions.}
  \label{fig:pipeline1}
\end{figure*}

\vspace{-5pt}
In the setting of human action-reaction synthesis, our primary goal is to generate the reaction $\bx_1=\{x_1^i\}_{i=1}^H$ conditioned on an arbitrary action $\bx_0 = \{x_0^i\}_{i=1}^H$ of length $H$. The condition $\bc$ can be action $\bx_0$, or it can be a signal such as an action label, text, audio to instruct the interaction, which is optional for intention-agnostic scenarios. 
We utilize SMPL-X~\cite{smplx} human model to represent the human motion sequence as~\citep{xu2024regennet} to improve the modeling of human-human interactions.
Thus, the reaction can be represented as $x^i_1 = \left[\bm{\theta}^{x_1}_i, \bm{q}^{x_1}_i, \bm{\gamma}^{x_1}_i\right]$
where $\bm{\theta}^{x_1}_i\in\mathbb{R}^{3K}$, $\bm{q}^{x_1}_i\in \mathbb{R}^3$, $\bm{\gamma}^{x_1}_i\in \mathbb{R}^3$ are the pose parameters, the global orientation, and the root translation of the person, respectively. Total number $K$ of body joints, including the jaw, eyeballs, and fingers, is 54.
The main pipeline of our \methodname~model is provided in Fig.~\ref{fig:pipeline1}. In this section, we first introduce the Human Action-Reaction Flow Matching in Sec.~\ref{sec:flow}.   
Then, we present our training-free physical constraint guidance method to address the issue of physically implausible human-human inter-penetrations in Sec.~\ref{sec:guidance}.

\subsection{Human Action-Reaction Flow Matching}
\label{sec:flow}
\paragraph{Flow Matching Overview.}
Given a set of samples from an unknown data distribution $q(\mathbf{x})$, the goal of flow maching is to learn a \textit{flow} that transforms a prior distribution $p_0(\mathbf{x})$ 
towards a target data distribution $p_1(\mathbf{x}) {\approx} q(\mathbf{x})$ along the probability path $p_t(\mathbf{x})$. 
The time-dependent flow $\phi_t(\bx)$ is defined by a vector field $\mathbf{v}(\mathbf{x},t): \mathbb{R}^d \times  [0, 1]  \rightarrow \mathbb{R}^d$ which establishes the flow through a neural ODE: 
\begin{align}\label{eq:ode}
    \frac{d}{dt}\phi_t(\mathbf{x}) = \mathbf{v}(\phi_t(\mathbf{x}), t), \quad \quad
    \phi_0(\mathbf{x}) &= \mathbf{x}~.
\end{align}
Given a predefined probability path $p_t(\mathbf{x})$ and a corresponding vector field $\mathbf{u}_t(\mathbf{x})$, one can regress the vector field $\mathbf{u}_t(\mathbf{x})$ with a neural network $\mathbf{v}_\theta(\mathbf{x}_{t},t)$ parameterized by $\theta$,  and the Flow Maching (FM) objective is as follows:
\begin{align}\label{eq:fm}
    \min_{\theta} \mathbb{E}_{t, p_t(\mathbf{x})} \| \mathbf{v}_\theta(\mathbf{x}_t,t) - \mathbf{u}_t(\mathbf{x}) \|^2.
\end{align} 
By defining the conditional probability path as a linear interpolation between $p_0$ and $p_1$, the intermediate process becomes: $\bx_t = t \bx_1 + [1 - (1 - \sigma_\text{min}) t] \bx_0$, where $\sigma_{\text{min}}>0$ is a small amount of noise.
Both training and sampling are simplified by fitting a linear trajectory in contrast to diffusion paths. When extra condition signals $\bc$ are required, they can be directly incorporated into the vector field estimator $\bv_\theta(\mathbf{x}_t, t)$ as $\mathbf{v}(\mathbf{x}_t, t, \bc)$.
Therefore, the training objective is as follows:
\begin{align}
\label{eqn:cfm}
    \min_{\theta} \mathbb{E}_{t, p(\mathbf{x}_0), q(\mathbf{x}_1)} \Big\| \mathbf{v}_\theta(\mathbf{x}_t, t, c) - \Bigl(\mathbf{x}_1 -  (1 - \sigma_{\text{min}}) \mathbf{x}_0 \Bigr) \Big\|^2.
\end{align} 
 
\paragraph{\textbf{Action-Reaction Flow Matching.}}
Different from previous diffusion-based methods\citep{xu2024regennet,tevet2022human_mdm,li2024interdance,agrol} that rely on cumbersome conditional mechanisms, 
we adopt Flow Matching to directly construct a mapping from action distribution to reaction distribution~(See Fig.~\ref{fig:colour}). Specially, we build a generative model $f$ parametrized by $\theta$ to synthesize the reaction $\bx_1 = f_\theta(\bx_0, \bc)$, given action $\bx_0$, instead of $\bx_1 = f_\theta(\bz, \bc)$ in diffusion, given a sampled Gaussian noise vector $\bz$.
Given the reaction $\bx_1$ sampled from the reaction distribution and the coupled action $\bx_0$ from the action distribution, the intermediate process can be written as
\begin{equation}\label{eqn:flow}
{
\bx_t = t \bx_1 + [1 - (1 - \sigma_\text{min}) t] \bx_0,
}
\end{equation}
where $t$ is the timestep, $\sigma_{\text{min}}>0$ is a small amount of noise. In our setting, the generative process is conditionally formulated as $p( \bx_{t_{n+1}}|\bx_{t_n},c)$. We use a neural 
network $G$ to directly predict the clean body poses, \emph{i.e.}, $\hat{\bx}_1=G_\theta(\bx_t, t, c)$, instead of predicting vector fields in previous works~\cite{hu2023motion,lipman2022flow}. This strategy is both straightforward and effective, since many geometric losses directly act on the predicted $\hat{\bx}_1$. 
We compared and analyzed the results of predicting vector fields ($\bv$-prediction) and clean body poses ($\bx_1$-prediction) in Sec.~\ref{sec:ablation}. Note that $\bx_1$ in flow maching usually corresponds to $\bx_0$ in previous literature on diffusion models. Depending on the specific application, $G$ can be implemented by Transformers~\citep{vaswani2017attention} or MLP networks.
The training objective of our flow model is as follows:
\begin{equation}\label{eqn:loss_diff}
{
~~~~\mathcal{L}_\text{fm} = \mathbb{E}_{\bx_1 \sim q(\bx_1), \bx_0 \sim p(\bx_0), t \sim [0,1]}[\| \bx_1 - G_\theta(\bx_t, t, c)\|_2^2].
}
\end{equation} 
Following~\citep{xu2024regennet}, we employ explicit interaction losses to evaluate the relative distances of body pose $\bm{\theta}(\bx_1 , \bx_0)$, orientation $\bm{q}(\bx_1 , \bx_0)$ and translation $\bm{\gamma}(\bx_1 , \bx_0)$ between the actor and reactor.
We use a forward kinematic function to transforms the rotation pose into joint positions for calculating $\bm{\theta}(\bx_1 , \bx_0)$ , and converts the rotation poses to rotation matrices for calculating $\bm{q}(\bx_1 , \bx_0)$. The interaction loss is defined as
\begin{equation} \label{eqn:loss_interact}
  \begin{aligned}
  ~~~~~~~~~~~&\mathcal{L}_\text{inter} = \frac{1}{H}\biggl( \| \bm{\theta}(\bx_1 , \bx_0) - \bm{\theta}(\hat{\bx}_1 , \bx_0)\|_{2}^{2} \\ &+  \| \bm{q}(\bx_1 , \bx_0) - \bm{q}(\hat{\bx}_1 , \bx_0)\|_{2}^{2} +  \| \bm{\gamma}(\bx_1 , \bx_0) - \bm{\gamma}(\hat{\bx}_1 , \bx_0)\|_{2}^{2}\biggr).
  \end{aligned}
\end{equation}
Our overall training loss is  $\mathcal{L}_\text{all} = \mathcal{L}_\text{fm} + \lambda_\text{inter}\cdot\mathcal{L}_\text{inter}$, and $\lambda_\text{inter}$ is the loss weight. 

\paragraph{\textbf{Sampling based on $\bx_1$-prediction.}} Since our neural network outputs $\hat{\bx}_1$, we require to construct an equivalent relationship between the neural network's predictions of $\bv$ and $\bx_1$.
The equivalent form of parameterization~Eq. \ref{eq:transform} derived from our appendix is as follows:
\begin{equation}\label{eqn:transform}
{
\bv_\theta(\mathbf{x}_t, t, c) = \frac{\hat{\bx}_{1} - (1 - \sigma_\text{min})\bx_t}{1 - (1 - \sigma_\text{min})t},~~~~~~~~~~~
}
\end{equation}
Then, our sampling based on $\bx_1$-prediction can be achieved by first sampling $\mathbf{x}_0$ and then solving Eq. \ref{eq:ode} employing an ODE solver~\citep{Runge, Kutta, alexander1990solving} through our trained neural network $\hat{\bx}_1=G_\theta(\bx_t, t, c)$. We use the Euler ODE solver and discretization process involves dividing the procedure into $N$ steps, leading to the following formulation:
\begin{equation}
\bx_{t_{n+1}} \leftarrow \bx_{t_n} + (t_{n+1}-t_{n})~\bv_{\theta}(\bx_{t_n}, t_n,\bc),
\label{eq:sample}
\end{equation}
where the integer time step $t_1 = 0 < t_2 < \cdots < t_N = 1$.
By using equivalent form of parameterization Eq. \ref{eqn:transform}, we finally obtain our flow maching sampling formulation based on $\bx_1$-prediction:
\vspace{-0pt}
\begin{align}
\label{eq:x1}
\bx_{t_{n+1}} \leftarrow \frac{1-(1 - \sigma_\text{min})t_{n+1}}{1-(1 - \sigma_\text{min})t_{n}} \bx_{t_{n}} + \frac{t_{n+1}-t_{n}}{1-(1 - \sigma_\text{min})t_{n}} ~\hat{\bx}_1,
\end{align}
which is more suitable for human motion generation. Detailed derivation is provided in \textbf{Appendix}~\ref{append:derivation}. 

\subsection{ARFlow Sampling with Physical Constraint Guidance} 
\label{sec:guidance}
There exist many physically implausible inter-penetrations between the actor and reactor in the generated results of current diffusion-based methods~\cite{xu2024regennet,tevet2022human_mdm,agrol}, owing to the sampling process that disregards physical constraints. 
To address the issue, we employ a penetration gradients $\nabla \mathcal{L}_\text{pene}$ to guide the sampling process. We follow~\cite{karunratanakul2024optimizing, li2024interdance} to calculate the signed distance function (SDF) between the actor and the reactor as our penetration loss functon:
\begin{equation}
    \mathcal{L}_\text{pene}(\bx) \defeq \sum_{i,h} -\mathrm{min} \Bigl( \mathrm{SDF}(\psi^h_i(\bx)), \zeta \Bigr),
\end{equation}
where $\psi^h_i(\bx)$ represents the position of the $i$-th joint in the $h$-th frame of the generated reaction motion $\bx$, the $\zeta$ defines the safe distance between the actor and the reactor, beyond which the gradient becomes zero, and $\mathrm{SDF}$ is the signed distance function for an actor in the $h$-th frame, which dynamically changes across frames.

Traditional diffusion guidance methods~\citep{chung2023diffusion,karunratanakul2023guided,tian2024gaze} first estimate $\hat{\bx}_1$ from current state $\bx_{t_{n}}$ with a denoiser network $\epsilon_\theta(\bx_{t_{n}}, t_{n},\bc)$, and then calculate gradients of the loss function with respect to current state $\bx_{t_{n}}$,
so it inevitably requires differentiation of the neural network, resulting in inaccurate gradients $\nabla \mathcal{L}_\text{pene}$.
\vspace{-0pt}
To avoid this issue, we directly updates the gradient at $\hat{\bx}_1$:
\begin{equation}
\hat{\bx}'_1 \leftarrow \hat{\bx}_1 - \lambda_\text{pene}\nabla_{\hat{\bx}_1} \mathcal{L}_\text{pene}(\hat{\bx}_1),
\label{eq:gradient}
\end{equation}
where $\hat{\bx}_1$ is the clean body poses predicted by our neural network $G_\theta$ and $\lambda_\text{pene}$ is the guidance strength. Then, we use the linear interpolation of Flow Matching to project back to the intermediate state of learned FM path.

\begin{figure}[h]
            \centering
            \begin{minipage}{0.5\textwidth}
			\centering
			\includegraphics[width=\textwidth]{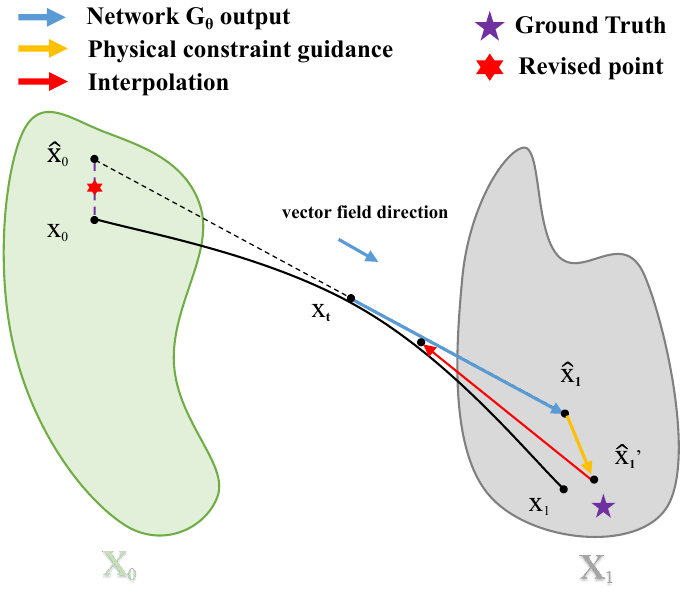} 
			\caption{Illustration of our ARFlow sampling with physical               constraint guidance.}
			\label{fig:projection}
		\end{minipage}
		\hfill
		\begin{minipage}{0.4\textwidth}
                \centering
                \footnotesize
                \resizebox{0.9\textwidth}{!}{
                    \begin{tabular}{lcccc}
                    \toprule
                     & FID $\downarrow$~ & \multicolumn{1}{p{1.7cm}}{\centering IF $\downarrow$ } & IV $\downarrow$\\ 
                     \midrule
                     Real & 0.09 & 21.96\% & 5.35 \\ 
                     \midrule
                     \multicolumn{4}{c}{~~~~~~~Offline} \\
                     \midrule
                    MDM  & 22.04  & 15.45\% & 3.36 & \\
                    ReGenNet  & 6.19  & 19.14\% & 3.07 & \\  
                    \methodname  & \textbf{5.00}  & \textbf{9.67\%} & \textbf{1.94} & \\  
                    \midrule
                    {MDM $^\text{vanilla}$}  & 20.42 & 13.29\% & 2.26  \\ 
                    {ReGenNet $^\text{vanilla}$}  & 7.69 & 17.85\% & 2.16 \\ 
                    {\methodname $^\text{vanilla}$} & 5.19 & 6.37\% & 0.31 \\
                    {\textbf{\methodname} $^\text{improved}$} & \textbf{5.05} & \textbf{3.73\%} & \textbf{0.23}\\
                    \midrule
                    \multicolumn{4}{c}{~~~~~~~Online} \\
                    \midrule
                    MDM  & 51.11  & 32.63\% & 17.97 & \\
                    ReGenNet  & 11.00  & 13.84\% & 3.50 & \\  
                    \methodname  & \textbf{7.89}  & \textbf{8.39\%} & \textbf{3.26} & \\  
                    \midrule
                    {MDM $^\text{vanilla}$}  & 40.03 & 22.63\% & 10.35 \\ 
                    {ReGenNet $^\text{vanilla}$}  & 11.63 & 13.05\% & 2.71 \\ 
                    {ARFlow $^\text{vanilla}$} & 8.61 & 3.29\% & 1.44 \\ 
                    {\textbf{\methodname} $^\text{improved}$} & \textbf{8.07} & \textbf{3.23\%} & \textbf{0.53}\\ 
                    \bottomrule
                    \end{tabular}
                }
                \captionof{table}{Human action-reaction synthesis with physical constraint guidance on NTU120-AS. \textbf{Bold} indicates the best result.}  \label{table:result_guidance}
		\end{minipage}
        \end{figure}

\noindent\textbf{Interpolation between action and reaction.} 
Since we build flow matching between the action and reaction distribution, unlike the process of adding noise back in diffusion, our interpolation process depends on the initial point $\bx_0$, so we need to project towards the direction of the initial point. The Flow Matching sampling algorithm Eq.~\ref{eq:x1} is equivalent to the following formulation:
\begin{align}\label{eqn:x0}
~~~~~~~~~~~~\hat{\bx}_0  \leftarrow~&\hat{\bx}_1 + \frac{\bx_{t_{n}} - (1 + \sigma_\text{min}t) \hat{\bx}_1}{1 - (1 - \sigma_\text{min})t_{n}},\\
\label{eqn:tradition}
\bx_{t_{n+1}} \leftarrow~&t_{n+1} \hat{\bx}_1 + [1 - (1 - \sigma_\text{min}) t_{n+1}] ~\hat{\bx}_0.
\end{align}
This sampling process essentially finds a $\hat{\bx}_0$ along the opposite direction of the current velocity field for linear interpolation as Eq.~\ref{eqn:x0} (black dotted lines in Fig.~\ref{fig:projection}). Obviously, this predicted $\hat{\bx}_0$ deviates from the initial point $\bx_0$ (purple dotted lines in Fig.~\ref{fig:projection}). In order to correct the projection direction of traditional Flow Matching, we use a weight factor $w$ to weight $\hat{\bx}_0$ and $\bx_0$:
\begin{equation}
\hat{\bx}^*_0 \leftarrow w \hat{\bx}_0 + (1-w)\bx_0.
\end{equation}
and use $\hat{\bx}^*_0$ as our final endpoint for interpolation. The ARFlow sampling and traditional sampling algorithm for $\bx_1$-prediction with physical constraint guidance are shown in Algorithm~\ref{alg:arflow} and \ref{alg:guidance_x} respectively. In practice, we use $\bx_1$-prediction for its better performance. Under iterative sampling and physical constraint guidance, our method can generate more realistic and physically-plausible reaction motions.

\noindent\textbf{Stochastic sampling to enhance diversity of reactions.}
To further enhance the diversity of reactions generated by ARFlow, we incorporate randomness into the sampling process. The interpolation~Eq.~\ref{eqn:tradition} can be written in the following equivalent form:
\begin{equation}
\bx_{t_{n+1}} \leftarrow \hat{\bx}_1  + (1 -  t_{n+1}) (\hat{\bx}_0-\hat{\bx}_1) +  \sigma_\text{min} t_{n+1} \hat{\bx}_0.
\end{equation}
The interpolation process can be understood as a projection in the opposite direction of the current learned velocity field $\hat{\bx}_1-{\hat{\bx}}_0$. 
Thus, we can weight the projection direction $\hat{\bx}^*_0 - \hat{\bx}'_1$
and stochastic direction $d_{\text{random}}$ to incorporate randomness:
\begin{align}
&d_{\text{mix}} \leftarrow \hat{\bx}^*_0 - \hat{\bx}'_1 + \beta [d_{\text{random}} - (\hat{\bx}^*_0 - \hat{\bx}'_1 )],\\
\label{eqn:random}
&\bx_{t_{n+1}} \leftarrow ~\hat{\bx}'_1  + (1 -  t_{n+1}) d_{\text{mix}} +  \sigma_\text{min} t_{n+1} \hat{\bx}^*_0,
\end{align}
where $\beta$ is the factor to control the strength of randomness. 
Our guidance method is actually a refined fine-tuning, which may be not suitable for training. 
In addition, the loss during the training mainly measures the difference between generated results and ground truth, while our guidance during the inference phase can provide more flexible guidance based on the quality of the generated results.

\noindent\textbf{Metrics.} To qualitatively measure the degree of penetration, we introduced two metrics:

1) \textbf{Intersection Volume (IV).} Penetrate in~\cite{yuan2023physdiff,han2024reindiffuse} just measures ground penetration which is not suitable for measuring the degree of penetration between humans. Interpenetration in~\cite{liu2024physreaction} can only be computed as rigid bodies in the physics simulation. Inspired by Solid Intersection Volume (IV)~\cite{zhou2022toch,liu2024geneoh}, we measure human-human inter-penetration by voxelizing actor and reactor meshes and reporting the volume of voxels occupied by both. Intersection Volume (IV) is defined as
\begin{equation}
\label{equ:if}
IV = \frac{1}{H\cdot N_\text{total}} \sum_{i=1}^{N_\text{total}}\sum_{h=1}^{H}V^h_\text{pene},   
\end{equation}
where $V^h_\text{pene}$ represents intersection volume of frame $h$ and $N_\text{total}$ denotes the total number of samples.  

2) \textbf{Intersection Frequency (IF).} Inspired by Contact Frequency in~\cite{li2024interdance,siyao2024duolando}, we introduce IF to measure the frequency of inter-penetration, \emph{i.e.}
\begin{equation}
\label{equ:if}
IF = f_\text{pene} / F_\text{total},
\end{equation}
where $f_\text{pene}$ represents the number of inter-penetration frames and $F_\text{total}$ is the total number of frames.


\newcommand{\comment}[1]{\textcolor{gray}{#1}}

\begin{figure}[ht]
\begin{minipage}[t]{0.48\textwidth}
\begin{algorithm}[H]
\small
\caption{
Sampling algorithm with vanilla guidance of physical constraints. 
}
\label{alg:guidance_x}
\begin{algorithmic}[1]
\STATE{\textbf{Input}: $\mathcal{L}_\text{pene}$ the loss function ; $G$ and $\theta$ the clean body poses predictor with pretrained parameters}
\STATE{\textbf{Parameters}: $N$ the number of sampling steps; 
$\lambda_\text{pene}$ the guidance strength}
\STATE{Sample $\bx_0$ from the action distribution}
\FOR{$n=1,2,...,N-1$}
\medskip
\STATE{$\hat{\bx}_1 \leftarrow G_\theta(\bx_{t_{n}}, t_{n},\bc) $}
\medskip
\STATE \comment{\small \# Flow Matching $\bx_1$-prediction \\
sampling (Eq. \ref{eq:x1})}
\STATE{$\bx'_{t_{n+1}} \leftarrow \frac{1-t_{n+1}}{1-t_{n}} \bx_{t_{n}}
+ \frac{t_{n+1}-t_{n}}{1-t_{n}} ~\hat{\bx}_1$} 
\medskip
\STATE \comment{\small \# Physical constraint guidance}
\STATE{$\bx_{t_{n+1}} \leftarrow \bx'_{t_{n+1}} - \lambda_\text{pene}\nabla_{\hat{\bx}_{t_{n}}} \mathcal{L}_\text{pene}(\hat{\bx}_1)$}
\smallskip
\ENDFOR
\STATE{\textbf{Return}: The reaction motion $\bx_{1} = \bx_{t_N}$}
\end{algorithmic}
\end{algorithm}
\end{minipage}
\hfill
\begin{minipage}[t]{0.48\textwidth}
\begin{algorithm}[H]
\small
\caption{
Sampling algorithm with improved guidance of physical constraints. 
}
\label{alg:arflow}
\begin{algorithmic}[1]
\STATE{\textbf{Input}: $\mathcal{L}_\text{pene}$ the loss function ; $G$ and $\theta$ the clean body poses predictor with pretrained parameters}
\STATE{\textbf{Parameters}: $N$ the number of sampling steps; 
$\lambda_\text{pene}$ the guidance strength; $w$ weight factor}
\STATE{Sample $\bx_0$ from the action distribution}
\FOR{$n=1,2,...,N-1$}
\STATE{$\hat{\bx}_1 \leftarrow G_\theta(\bx_{t_{n}}, t_{n},\bc) $}
\STATE{$\hat{\bx}_0 \leftarrow \hat{\bx}_1 + \frac{\bx_{t_{n}} - (1 + \sigma_\text{min}t) \hat{\bx}_1}{1 - (1 - \sigma_\text{min})t_{n}}$} ~~~~~~\comment{\small \#~ (Eq. \ref{eqn:x0})}
\STATE \comment{\small \# Physical constraint guidance at $\hat{\bx}_1$(Eq. \ref{eq:gradient})}
\STATE{$\hat{\bx}'_1 \leftarrow \hat{\bx}_1 - \lambda_\text{pene}\nabla_{\hat{\bx}_1} \mathcal{L}_\text{pene}(\hat{\bx}_1)$}
\STATE \comment{\small \# Direction correction}
\STATE{$\hat{\bx}^*_0 \leftarrow w \hat{\bx}_0 + (1-w)\bx_0$} 
\STATE \comment{\small \# Interpolation (Eq. \ref{eqn:tradition})}
\STATE{$\bx_{t_{n+1}} \leftarrow t_{n+1} \hat{\bx}'_1 + [1 - (1 - \sigma_\text{min}) t_{n+1}] ~\hat{\bx}^*_0$}
\ENDFOR
\STATE{\textbf{Return}: The reaction motion $\bx_{1} = \bx_{t_N}$}
\end{algorithmic}
\end{algorithm}
\end{minipage}
\end{figure}
\section{Experiments}
\vspace{-2mm}
\begin{figure*}[h]
  \centering
  \includegraphics[width=0.8\linewidth]{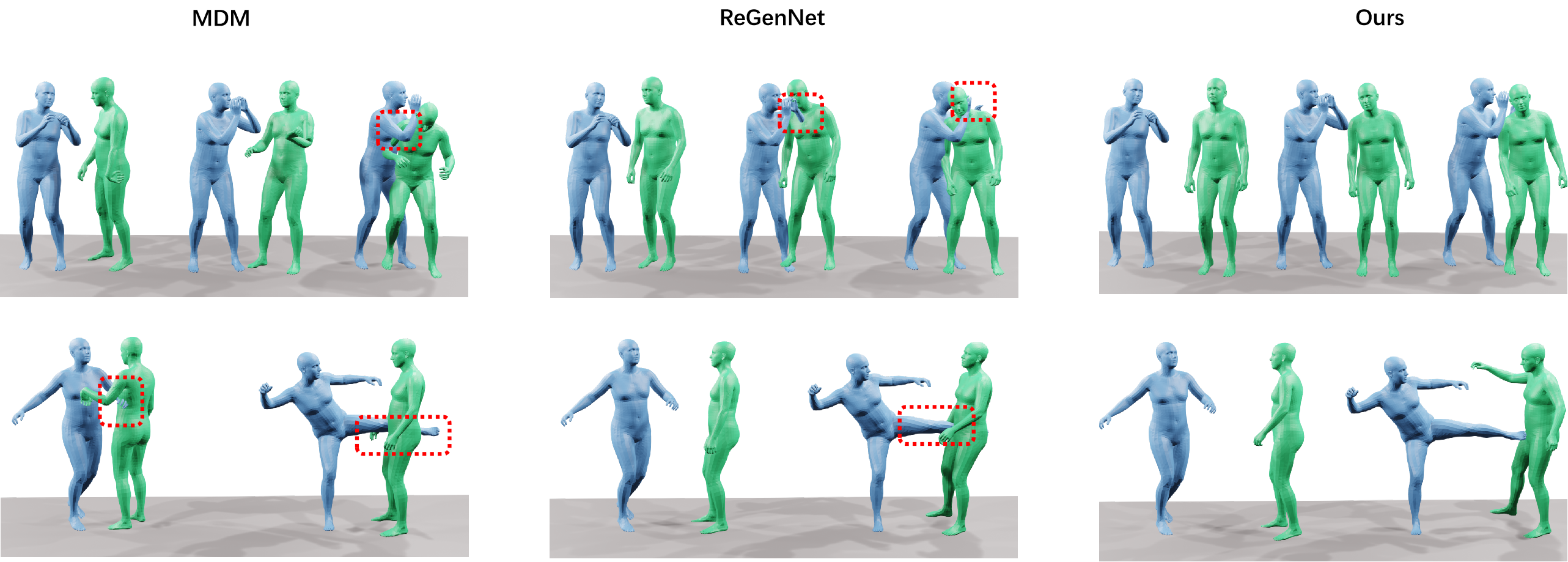}
  \caption{\textbf{Qualitative comparisons} of human action-reaction synthesis results. \textcolor{blue}{Blue} for actors and \textcolor{green}{Green} for reactors.}
  \label{fig:main}
  \vspace{-6mm}
\end{figure*}

Our experiment setting of human action-reaction synthesis is \textbf{online} and \textbf{unconstrained} as in ~\citep{xu2024regennet} for its significant potential for practical applications.
\textbf{Online} represents real-time reaction generation where future motions of the actor are not visible to the reactor, and the opposite is \textbf{offline} to relax the synchronicity. \textbf{Unconstrained} means that the intention of the actor is invisible to the reactor. 
To demonstrate the universality of our flow matching and physical guidance methods, we also conducted offline setting experiments.

\begin{table*}[t]
  \begin{center}
  \resizebox{0.7\textwidth}{!}{
  \begin{tabular}{l c c c c }
  \toprule
  Method & FID$\downarrow$ & Acc.$\uparrow$ & Div.$\rightarrow$ & Multimod.$\rightarrow$ \\
  \midrule
  Real & $0.09^{\pm0.00}$ & $0.867^{\pm0.0002}$ & $13.06^{\pm0.09}$ & $25.03^{\pm0.23}$ \\
  \midrule
  cVAE~\cite{vae} & $70.10^{\pm3.42}$ & $ 0.724^{\pm0.0002}$ & $11.14^{\pm0.04}$ & $18.40^{\pm0.26}$\\
  AGRoL~\cite{agrol} & $44.94^{\pm2.46}$ & $0.680^{\pm0.0001}$ & $ 12.51^{\pm0.09}$ & $19.73^{\pm0.17}$ \\
  MDM~\cite{tevet2022human_mdm} & $54.54^{\pm3.94}$ & $0.704^{\pm0.0003}$ & $11.98^{\pm0.07}$ & $19.45^{\pm0.20}$ \\
  MDM-GRU~\cite{tevet2022human_mdm} & $24.25^{\pm1.39}$ & $0.720^{\pm0.0002}$ & $\mathbf{13.43^{\pm0.09}}$ & $22.24^{\pm0.29}$\\
  ReGenNet~\cite{xu2024regennet} & $\underline{11.00^{\pm0.74}}$ & $\mathbf{0.749^{\pm0.0002}}$ & $13.80^{\pm0.16}$ & $\underline{22.90^{\pm0.14}}$ \\
  \textbf{\methodname} & $\mathbf{7.89^{\pm0.18}}$ & $\underline{0.743^{\pm0.0002}}$ & $ \underline{13.60^{\pm0.10}}$ & $\mathbf{24.11^{\pm0.13}}$ \\
  \bottomrule
  \end{tabular}}
  \end{center}
  \vspace{-3mm}
  \caption[caption]{\textbf{Comparison to state-of-the-art} on the \textit{online, unconstrained} setting on NTU120-AS. $\rightarrow$ denotes that the result closer to Real is better, and $\pm$ represents 95\% confidence interval. We highlight the best result in \textbf{Bold} and the second best in \underline{underline}.}
  \label{tab:cmdm}
  \vspace{-2mm}
\end{table*}

\begin{table*}[t]
\vspace{-1mm}
\begin{center}
\resizebox{0.7\textwidth}{!}{
\begin{tabular}{l c c c c c c c }
\toprule
Method & FID$\downarrow$ & Acc.$\uparrow$ & Div.$\rightarrow$ & Multimod.$\rightarrow$ \\
\midrule
Real & $0.75^{\pm0.18}$ & $0.691^{\pm0.0093}$ & $7.15^{\pm1.27}$ & $12.94^{\pm0.96}$ \\
\midrule
cVAE~\cite{vae} & $17.33^{\pm17.14}$ & $0.552^{\pm0.0024}$ & $8.20^{\pm0.57}$ & $11.44^{\pm0.35}$ \\
AGRoL~\cite{agrol} & $64.83^{\pm277.8}$ & $\underline{0.644^{\pm0.0039}}$ & $\mathbf{7.00^{\pm0.95}}$ & $11.33^{\pm0.65}$ \\
MDM~\cite{tevet2022human_mdm} & $18.40^{\pm7.95}$ & $\mathbf{0.647^{\pm0.0035}}$ & $5.89^{\pm0.33}$ & $10.96^{\pm0.27}$ \\
MDM-GRU~\cite{tevet2022human_mdm} & $18.63^{\pm25.87}$ & $0.574^{\pm0.0046}$ & $6.20^{\pm0.24}$ & $10.49^{\pm0.32}$ \\
ReGenNet~\cite{xu2024regennet} & $\underline{13.76^{\pm4.78}}$ & $0.601^{\pm0.0040}$ & $6.35^{\pm0.24}$ & $\underline{12.02^{\pm0.33}}$ \\
\textbf{\methodname} & $\mathbf{10.92^{\pm3.70}}$ & $0.600^{\pm0.0040}$ & $\underline{6.68^{\pm0.25}}$ & $\mathbf{12.74^{\pm0.17}}$ \\
\bottomrule
\end{tabular}}
\end{center}
\vspace{-2mm}
\caption[caption]{\textbf{Comparison to state-of-the-art } on the \textit{online, unconstrained} setting on Chi3D-AS. $\rightarrow$ denotes that the result closer to Real is better, and $\pm$ represents 95\% confidence interval. We highlight the best result in \textbf{Bold} and the second best in \underline{underline}.}
\label{tab:cmdm_chi3d}
\vspace{-6mm}
\end{table*}

\begin{table*}[htbp]
  \vspace{-1mm}
  \begin{center}
  \resizebox{0.7\textwidth}{!}{
  \begin{tabular}{l l c c c c c }
  \toprule
  \multirow{2}{*}{Class} & \multirow{2}{*}{Settings}  & \multirow{2}{*}{FID$\downarrow$} & \multirow{2}{*}{Acc.$\uparrow$} & \multirow{2}{*}{Div.$\rightarrow$} & \multirow{2}{*}{Multimod.$\rightarrow$}  & \multirow{2}{*}{Latency(ms)}\\
  & \\
  \midrule
  & Real & $0.085^{\pm0.0003}$ & $0.867^{\pm0.0002}$ & $13.063^{\pm0.0908}$ & $25.032^{\pm0.2332}$ & -\\
  \midrule
  \multirow{2}{*}{Prediction} & 1) $x_1$  &$\mathbf{7.894^{\pm0.1814}}$ & $\mathbf{0.743^{\pm0.0002}}$ & $\mathbf{13.599^{\pm0.1005}}$ & $\mathbf{24.105^{\pm0.1310}}$ & - \\
  &2) $v$ & $14.726^{\pm0.2143}$ & $0.743^{\pm0.0002}$ & $14.154^{\pm0.0923}$ & $23.329^{\pm0.1125}$ & - \\
  \midrule
  \multirow{1}{*}{Guidance} & w. $\mathcal{L}_\text{pene}$ & $8.073^{\pm0.1981}$ & $0.741^{\pm0.0002}$ & $13.713^{\pm0.1079}$ & $24.077^{\pm0.1370}$ & - \\
  \midrule
  \multirow{4}{*}{Timesteps} & 2 & $15.965^{\pm0.2728}$ & $0.733^{\pm0.0002}$ & $13.740^{\pm0.0896}$ & $26.767^{\pm0.1440}$ & $0.055$\\
  & 5  &$\mathbf{7.894^{\pm0.1814}}$ & $0.743^{\pm0.0002}$ & $\mathbf{13.599^{\pm0.1005}}$ & $\mathbf{24.105^{\pm0.1310}}$ & $0.111$\\
  & 10 & $8.273^{\pm0.3862}$ & $0.721^{\pm0.0002}$ & $14.108^{\pm0.0779}$ & $22.995^{\pm0.1274}$ & $0.232$\\
  & 100 & $8.259^{\pm0.3902}$ & $0.747^{\pm0.0002}$ & $14.173^{\pm0.1024}$ & $23.619^{\pm0.1214}$ & $2.273$\\
  \midrule
  Best &\textbf{\methodname} & $\mathbf{7.894^{\pm0.1814}}$ & $0.743^{\pm0.0002}$ & $ \mathbf{13.599^{\pm0.1005}}$ & $\mathbf{24.105^{\pm0.1310}}$ & $0.111$\\
  \bottomrule
  \end{tabular}
  }
  \end{center}
  \vspace{-2mm}
  \caption[caption]{\textbf{Ablation studies} on the \textit{online, unconstrained} setting on the NTU120-AS dataset. \textbf{Bold} indicates the best result in our method.}
  \label{tab:ablations_online}
  \end{table*}

\vspace{-2mm}
\subsection{Experiment setup}
\vspace{-2mm}
\noindent\textbf{Evaluation Metrics.} 1) We adopt the following metrics to quantitatively evaluate results: Frechet Inception Distance (FID),~Action Recognition Accuracy (Acc.),~Diversity (Div.) and Multi-modality (Multimod.). 
For all these metrics widely used in previous human motion generation~\cite{action2motion,petrovich2021action,tevet2022human_mdm,xu2024regennet}, we use the action recognition model~\cite{stgcn} to extract motion features for calculating these metrics as in~\cite{xu2024regennet}. 
We generate 1,000 reaction samples by sampling actor motions from test sets and evaluate each method 20 times using different random seeds to calculate the average with the 95\% confidence interval as prior works~\cite{action2motion,petrovich2021action,tevet2022human_mdm,xu2024regennet}.
2) \textbf{Intersection Volume (IV)} measures human-human inter-penetration by voxelizing actor and reactor meshes and reporting the volume of voxels occupied by both.  3) \textbf{Intersection Frequency (IF)} measures the frequency of inter-penetration. More details about these metrics are provided in the supplementary.

\noindent\textbf{Datasets.} We evaluate our model on NTU120-AS, Chi3D-AS and InterHuman-AS datasets with SMPL-X~\cite{smplx} body models and actor-reactor annotations as in~\cite{petrovich2021action}.They contains 8118, 373 and 6022 human interaction sequences, respectively. \textquotedblleft AS\textquotedblright\cite{xu2024regennet} represents that they are an extended version of the original dataset~\cite{chi3d,nturgbd120,ntu-x,intergen}, which adds annotations to distinguish actor-reactor order of each interaction sequence and  SMPL-X body models for more detailed representations.
We adopt the 6D rotation representation~\cite{6d_rot} in all our experiments. 

\vspace{-6mm}
\subsection{Comparison to baselines}
\vspace{-2mm}

To evaluate the performance of our method, we adopt following baselines: 1) cVAE~\cite{vae}, commonly utilized in earlier generative models for human interactions; 2) MDM~\cite{tevet2022human_mdm}, the state-of-the-art diffusion-based method for human motion generation, and its variant MDM-GRU~\cite{tevet2022human_mdm}, which incorporates a GRU~\cite{cho2014learning} backbone; 3) AGRoL~\cite{agrol}, the current state-of-the-art method to generate full-body motions from sparse tracking signals, which adopts diffusion models with MLPs architectures; 4) ReGenNet~\citep{xu2024regennet}, the state-of-the-art diffusion-based method for human action-reaction synthesis on online, unconstraint setting as ours. Results are taken from tables of ReGenNet~\citep{xu2024regennet} where all methods use 5-timestep sampling.

For the NTU120-AS dataset in Tab.~\ref{tab:cmdm} and Chi3D-AS dataset in Tab.~\ref{tab:cmdm_chi3d}, our proposed \textbf{\methodname} notably outperforms baselines in terms of the FID metric,  demonstrating that our method better models the mapping between the action and the reaction distribution. 
Our method achieves the best FID and multi-modality, second best for the action recognition accuracy and diversity on the NTU120-AS dataset and the best FID and multi-modality, second best for the diversity on the Chi3D-AS dataset.
For a fair comparison, we use the pre-trained action recognition model in~\cite{xu2024regennet}, so our action recognition accuracy is very close to the results of~\cite{xu2024regennet}.
Given the restricted size of the Chi3D-AS test set, some fluctuations in the experimental results are to be expected. The results of InterHuman-AS dataset in Appendix~\ref{tab:cmdm_interhuman}
show our method also yields the best results compared to baselines. As for the \textbf{offline} setting, we replace the Transformer decoder units equipped with attention masks with an 8-layer Transformer encoder architecture just like ReGenNet~\cite{xu2024regennet}. As demonstrated in~Tab.~\ref{tab:cmdm_offline}, our model also achieves superior performance on most of the metrics.

\vspace{-2mm}
\subsection{Physical Constraint Guidance}
\vspace{-1mm}

\noindent{\textbf{Qualitative results.}}
In~Fig.~\ref{fig:main}, 
for the same action sequences sampled from the test sets of NTU120-AS datasets, visualization results demonstrate that both MDM and ReGenNet produce varying degrees of penetration between the actor and the reactor. In contrast, our method not only produces more physically plausible reactions, \emph{i.e.}, mitigate penetration problems, but also more responsive reactions, \emph{i.e.}, move backward when kicked, due to stronger modeling ability for causal relationship between actions and reactions.
For more visualizations and \textbf{videos} of the generated human reactions, please refer to the supplementary materials.

\noindent{\textbf{Quantitative results.}}
In~Tab.~\ref{table:result_guidance}, our \methodname~with physical constraint guidance achieves the lowest Intersection Volume, Intersection Frequency and FID than other baselines, which shows that our method achieves the lowest level of penetration while ensuring the highest generation quality. 
In addition, the performance is further improved through our ARFlow sampling with improved physical constraint guidance, which demonstrates the effectiveness of our guidance method. 
Although our guidance method effectively suppresses penetration, it also leads to a slight increase in FID, as FID only measures the similarity between generated results and the ground truth distribution and the dataset itself exists a certain degree of penetration.

\vspace{-2mm}
\subsection{Ablation Study}
\label{sec:ablation}
\noindent{\textbf{Network Prediction.}}
As depicted in~Tab.~\ref{sec:flow}, a straightforward and effective strategy is to estimate clean body poses directly through a neural network, \emph{i.e.}, $\bx_1$-prediction. We compared it with $\bv$-prediction and the results are listed on the Prediction setting in~Tab.~\ref{tab:ablations_online} and ~Tab.~\ref{tab:ablations_offline}. Obviously, $\bx_1$-prediction has demonstrated superior performance across both settings. The reason we analyze it is that the geometric losses to regularize the generative network during the training phase directly acts on the predicted clean body poses, while $\bv$-prediction requires using the predicted vector field to estimate the clean poses, so the models trained by $\bx_1$-prediction will achieve stronger overall performance.

\noindent{\textbf{Physical constraint guidance.}}
As we discussed earlier, the results in~Tab.~\ref{table:result_guidance} indicate that our physical constraint guidance can effectively suppress penetration occurrences, and it may also lead to a decrease in some other metrics shown in~Tab.~\ref{tab:ablations_online} and~Tab.~\ref{tab:ablations_offline}.
We also provide a qualitative comparison of the effects before and after using our physical constraint guidance
in~Tab.~\ref{fig:vis}. The qualitative and quantitative results demonstrated that our method achieves the lowest penetration level while maintaining the best quality of generated reactions.

\noindent{\textbf{Number of Euler sampling timesteps.}}
We present comprehensive evaluation results in both online and offline scenarios, with varying Euler sampling intervals (2, 5, 10 and 100 timesteps), including the latency of reaction generation per frame on online settings and overall latency on offline settings. The experimental results, as detailed in~Tab.~\ref{tab:ablations_online} and~Tab.~\ref{tab:ablations_offline}, suggest that the 5-timestep Euler sampling consistently achieves optimal performance, demonstrating superior FID scores while maintaining low latency across both evaluation settings. Thus, we adopt the 5-timestep inference  as the standard configuration like ~\citep{xu2024regennet} for all the experimental results reported in this study.

\vspace{-3mm}
\section{Conclusion}
\label{sec:conclution}
\vspace{-2mm}
In this work, we have presented Action-Reaction Flow Matching (ARFlow), a novel framework for human action-reaction synthesis that addresses the limitations of existing diffusion-based approaches. By establishing direct action-to-reaction mappings through flow matching, ARFlow eliminates the need for complex conditional mechanisms and computationally expensive noise-to-data transformations. ARFlow involves a reprojection sampling algorithm with physical constraint guidance to enable efficient, physically plausible motion generation while preventing body penetration artifacts. Extensive evaluations on the NTU120, Chi3D and InterHuman datasets demonstrate that ARFlow excels over existing methods, showing superior performance in terms of Fréchet Inception Distance and motion diversity. Additionally, it significantly reduces body collisions, as evidenced by our new Intersection Volume and Intersection Frequency metrics. 

\noindent{\textbf{Limitations.}}
Although we attempt to use a reprojection method to address the issue of manifold distortions—deviations from the natural motion distribution established by flow matching, this problem still exists and our penetration loss function is just a simple design. Addressing these challenge opens a promising avenue for future theoretical research, focusing on developing guidance mechanisms that ensure physical plausibility without compromising motion authenticity. 

\section*{Acknowledgement}
This work was supported by National Natural Science Foundation of China (No.62406195, No.62303319), Shanghai Local College Capacity Building Program (23010503100), ShanghaiTech AI4S Initiative SHTAI4S202404, HPC Platform of ShanghaiTech University, Core Facility Platform of Computer Science and Communication of ShanghaiTech University, and MoE Key Laboratory of Intelligent Perception and Human-Machine Collaboration (ShanghaiTech University) and Shanghai Engineering Research Center of Intelligent Vision and Imaging.
\bibliography{main}

\begin{thebibliography}{83}
\providecommand{\natexlab}[1]{#1}
\providecommand{\url}[1]{\texttt{#1}}
\expandafter\ifx\csname urlstyle\endcsname\relax
  \providecommand{\doi}[1]{doi: #1}\else
  \providecommand{\doi}{doi: \begingroup \urlstyle{rm}\Url}\fi

\bibitem[Albergo and Vanden-Eijnden(2023)]{albergo2022building}
Michael~S Albergo and Eric Vanden-Eijnden.
\newblock Building normalizing flows with stochastic interpolants.
\newblock In \emph{ICLR}, 2023.

\bibitem[Alexander(1990)]{alexander1990solving}
Roger Alexander.
\newblock Solving ordinary differential equations i: Nonstiff problems (e. hairer, sp norsett, and g. wanner).
\newblock \emph{Siam Review}, 1990.

\bibitem[Aliakbarian et~al.(2022)Aliakbarian, Cameron, Bogo, Fitzgibbon, and Cashman]{aliakbarian2022flag}
Sadegh Aliakbarian, Pashmina Cameron, Federica Bogo, Andrew Fitzgibbon, and Thomas~J Cashman.
\newblock Flag: Flow-based 3d avatar generation from sparse observations.
\newblock In \emph{CVPR}, pages 13253--13262, 2022.

\bibitem[Ao et~al.(2022)Ao, Gao, Lou, Chen, and Liu]{ao2022rhythmic}
Tenglong Ao, Qingzhe Gao, Yuke Lou, Baoquan Chen, and Libin Liu.
\newblock Rhythmic gesticulator: Rhythm-aware co-speech gesture synthesis with hierarchical neural embeddings.
\newblock \emph{TOG}, 41\penalty0 (6):\penalty0 1--19, 2022.

\bibitem[Aram~Davtyan and Favaro(2023)]{video_fm}
Sepehr~Sameni Aram~Davtyan and Paolo Favaro.
\newblock Efficient video prediction via sparsely conditioned flow matching.
\newblock In \emph{ICCV}, 2023.

\bibitem[Cervantes et~al.(2022)Cervantes, Sekikawa, Sato, and Shinoda]{cervantes2022implicit}
Pablo Cervantes, Yusuke Sekikawa, Ikuro Sato, and Koichi Shinoda.
\newblock Implicit neural representations for variable length human motion generation.
\newblock In \emph{ECCV}, pages 356--372. Springer, 2022.

\bibitem[Chen et~al.(2018)Chen, Rubanova, Bettencourt, and Duvenaud]{chen2018neural}
Ricky~TQ Chen, Yulia Rubanova, Jesse Bettencourt, and David~K Duvenaud.
\newblock Neural ordinary differential equations.
\newblock \emph{Advances in neural information processing systems}, 31, 2018.

\bibitem[Chen et~al.(2023)Chen, Jiang, Liu, Huang, Fu, Chen, and Yu]{chen2023executing_mld}
Xin Chen, Biao Jiang, Wen Liu, Zilong Huang, Bin Fu, Tao Chen, and Gang Yu.
\newblock Executing your commands via motion diffusion in latent space.
\newblock In \emph{CVPR}, 2023.

\bibitem[Cho et~al.(2014)Cho, Van~Merri{\"e}nboer, Gulcehre, Bahdanau, Bougares, Schwenk, and Bengio]{cho2014learning}
Kyunghyun Cho, Bart Van~Merri{\"e}nboer, Caglar Gulcehre, Dzmitry Bahdanau, Fethi Bougares, Holger Schwenk, and Yoshua Bengio.
\newblock Learning phrase representations using rnn encoder-decoder for statistical machine translation.
\newblock \emph{arXiv preprint arXiv:1406.1078}, 2014.

\bibitem[Chopin et~al.(2023)Chopin, Tang, Otberdout, Daoudi, and Sebe]{chopin2023interaction}
Baptiste Chopin, Hao Tang, Naima Otberdout, Mohamed Daoudi, and Nicu Sebe.
\newblock Interaction transformer for human reaction generation.
\newblock \emph{IEEE Transactions on Multimedia}, 2023.

\bibitem[Chung et~al.(2022)Chung, Sim, Ryu, and Ye]{chung2022improving}
Hyungjin Chung, Byeongsu Sim, Dohoon Ryu, and Jong~Chul Ye.
\newblock Improving diffusion models for inverse problems using manifold constraints.
\newblock \emph{Advances in Neural Information Processing Systems}, 35:\penalty0 25683--25696, 2022.

\bibitem[Chung et~al.(2023{\natexlab{a}})Chung, Kim, McCann, Klasky, and Ye]{chung2023diffusion}
Hyungjin Chung, Jeongsol Kim, Michael~T McCann, Marc~L Klasky, and Jong~Chul Ye.
\newblock Diffusion posterior sampling for general noisy inverse problems.
\newblock In \emph{11th International Conference on Learning Representations, ICLR 2023}, 2023{\natexlab{a}}.

\bibitem[Chung et~al.(2023{\natexlab{b}})Chung, Kim, Mccann, Klasky, and Ye]{chungdiffusion}
Hyungjin Chung, Jeongsol Kim, Michael~Thompson Mccann, Marc~Louis Klasky, and Jong~Chul Ye.
\newblock Diffusion posterior sampling for general noisy inverse problems.
\newblock In \emph{The Eleventh International Conference on Learning Representations}, 2023{\natexlab{b}}.

\bibitem[Dabral et~al.(2022)Dabral, Mughal, Golyanik, and Theobalt]{mofusion}
Rishabh Dabral, Muhammad~Hamza Mughal, Vladislav Golyanik, and Christian Theobalt.
\newblock Mofusion: A framework for denoising-diffusion-based motion synthesis.
\newblock \emph{arXiv preprint arXiv:2212.04495}, 2022.

\bibitem[Du et~al.(2023)Du, Kips, Pumarola, Starke, Thabet, and Sanakoyeu]{agrol}
Yuming Du, Robin Kips, Albert Pumarola, Sebastian Starke, Ali Thabet, and Artsiom Sanakoyeu.
\newblock Avatars grow legs: Generating smooth human motion from sparse tracking inputs with diffusion model.
\newblock \emph{arXiv preprint arXiv:2304.08577}, 2023.

\bibitem[Esser et~al.(2024)Esser, Kulal, Blattmann, Entezari, M{\"u}ller, Saini, Levi, Lorenz, Sauer, Boesel, et~al.]{esser2024scaling}
Patrick Esser, Sumith Kulal, Andreas Blattmann, Rahim Entezari, Jonas M{\"u}ller, Harry Saini, Yam Levi, Dominik Lorenz, Axel Sauer, Frederic Boesel, et~al.
\newblock Scaling rectified flow transformers for high-resolution image synthesis.
\newblock In \emph{Forty-first international conference on machine learning}, 2024.

\bibitem[Feng et~al.(2025)Feng, Wu, Yu, Deng, and Hu]{feng2025guidance}
Ruiqi Feng, Tailin Wu, Chenglei Yu, Wenhao Deng, and Peiyan Hu.
\newblock On the guidance of flow matching.
\newblock \emph{arXiv preprint arXiv:2502.02150}, 2025.

\bibitem[Fieraru et~al.(2020)Fieraru, Zanfir, Oneata, Popa, Olaru, and Sminchisescu]{chi3d}
Mihai Fieraru, Mihai Zanfir, Elisabeta Oneata, Alin-Ionut Popa, Vlad Olaru, and Cristian Sminchisescu.
\newblock Three-dimensional reconstruction of human interactions.
\newblock In \emph{CVPR}, pages 7214--7223, 2020.

\bibitem[Guo et~al.(2020)Guo, Zuo, Wang, Zou, Sun, Deng, Gong, and Cheng]{action2motion}
Chuan Guo, Xinxin Zuo, Sen Wang, Shihao Zou, Qingyao Sun, Annan Deng, Minglun Gong, and Li Cheng.
\newblock Action2motion: Conditioned generation of 3d human motions.
\newblock In \emph{{ACM} Multimedia}, pages 2021--2029. {ACM}, 2020.

\bibitem[Guo et~al.(2022{\natexlab{a}})Guo, Zou, Zuo, Wang, Ji, Li, and Cheng]{guo2022generating}
Chuan Guo, Shihao Zou, Xinxin Zuo, Sen Wang, Wei Ji, Xingyu Li, and Li Cheng.
\newblock Generating diverse and natural 3d human motions from text.
\newblock In \emph{Proceedings of the IEEE/CVF conference on computer vision and pattern recognition}, pages 5152--5161, 2022{\natexlab{a}}.

\bibitem[Guo et~al.(2022{\natexlab{b}})Guo, Zuo, Wang, and Cheng]{guo2022tm2t}
Chuan Guo, Xinxin Zuo, Sen Wang, and Li Cheng.
\newblock Tm2t: Stochastic and tokenized modeling for the reciprocal generation of 3d human motions and texts.
\newblock In \emph{ECCV}, 2022{\natexlab{b}}.

\bibitem[Habibie et~al.(2022)Habibie, Elgharib, Sarkar, Abdullah, Nyatsanga, Neff, and Theobalt]{habibie2022motion}
Ikhsanul Habibie, Mohamed Elgharib, Kripasindhu Sarkar, Ahsan Abdullah, Simbarashe Nyatsanga, Michael Neff, and Christian Theobalt.
\newblock A motion matching-based framework for controllable gesture synthesis from speech.
\newblock In \emph{ACM SIGGRAPH 2022 Conference Proceedings}, pages 1--9, 2022.

\bibitem[Han et~al.(2024)Han, Liang, Tang, Cheng, Liu, and Huang]{han2024reindiffuse}
Gaoge Han, Mingjiang Liang, Jinglei Tang, Yongkang Cheng, Wei Liu, and Shaoli Huang.
\newblock Reindiffuse: Crafting physically plausible motions with reinforced diffusion model.
\newblock \emph{arXiv preprint arXiv:2410.07296}, 2024.

\bibitem[Heusel et~al.(2017)Heusel, Ramsauer, Unterthiner, Nessler, and Hochreiter]{fid}
Martin Heusel, Hubert Ramsauer, Thomas Unterthiner, Bernhard Nessler, and Sepp Hochreiter.
\newblock Gans trained by a two time-scale update rule converge to a local nash equilibrium.
\newblock In \emph{{NIPS}}, pages 6626--6637, 2017.

\bibitem[Ho and Salimans(2021)]{ho2021classifier}
Jonathan Ho and Tim Salimans.
\newblock Classifier-free diffusion guidance.
\newblock In \emph{NeurIPS Workshop}, 2021.

\bibitem[Hoyet et~al.(2012)Hoyet, McDonnell, and O'Sullivan]{hoyet2012push}
Ludovic Hoyet, Rachel McDonnell, and Carol O'Sullivan.
\newblock Push it real: Perceiving causality in virtual interactions.
\newblock \emph{ACM Transactions on Graphics (TOG)}, 31\penalty0 (4):\penalty0 1--9, 2012.

\bibitem[Hu et~al.(2023)Hu, Yin, Ma, Chen, Fernando, Asano, Gavves, Mettes, Ommer, and Snoek]{hu2023motion}
Vincent~Tao Hu, Wenzhe Yin, Pingchuan Ma, Yunlu Chen, Basura Fernando, Yuki~M Asano, Efstratios Gavves, Pascal Mettes, Bjorn Ommer, and Cees~GM Snoek.
\newblock Motion flow matching for human motion synthesis and editing.
\newblock \emph{arXiv preprint arXiv:2312.08895}, 2023.

\bibitem[Javed et~al.(2024)Javed, Guo, Cheng, and Li]{javed2024intermask}
Muhammad~Gohar Javed, Chuan Guo, Li Cheng, and Xingyu Li.
\newblock Intermask: 3d human interaction generation via collaborative masked modelling.
\newblock \emph{arXiv preprint arXiv:2410.10010}, 2024.

\bibitem[Jiang et~al.(2023)Jiang, Chen, Liu, Yu, Yu, and Chen]{jiang2023motiongpt}
Biao Jiang, Xin Chen, Wen Liu, Jingyi Yu, Gang Yu, and Tao Chen.
\newblock Motiongpt: Human motion as a foreign language.
\newblock In \emph{NeurIPS}, 2023.

\bibitem[Jiang et~al.(2024)Jiang, Xiao, Lin, Zhang, Ren, Gao, Lin, Cai, Yang, and Liu]{jiang2024solami}
Jianping Jiang, Weiye Xiao, Zhengyu Lin, Huaizhong Zhang, Tianxiang Ren, Yang Gao, Zhiqian Lin, Zhongang Cai, Lei Yang, and Ziwei Liu.
\newblock Solami: Social vision-language-action modeling for immersive interaction with 3d autonomous characters.
\newblock \emph{arXiv preprint arXiv:2412.00174}, 2024.

\bibitem[Karunratanakul et~al.(2023)Karunratanakul, Preechakul, Suwajanakorn, and Tang]{karunratanakul2023guided}
Korrawe Karunratanakul, Konpat Preechakul, Supasorn Suwajanakorn, and Siyu Tang.
\newblock Guided motion diffusion for controllable human motion synthesis.
\newblock In \emph{Proceedings of the IEEE/CVF International Conference on Computer Vision}, pages 2151--2162, 2023.

\bibitem[Karunratanakul et~al.(2024)Karunratanakul, Preechakul, Aksan, Beeler, Suwajanakorn, and Tang]{karunratanakul2024optimizing}
Korrawe Karunratanakul, Konpat Preechakul, Emre Aksan, Thabo Beeler, Supasorn Suwajanakorn, and Siyu Tang.
\newblock Optimizing diffusion noise can serve as universal motion priors.
\newblock In \emph{Proceedings of the IEEE/CVF Conference on Computer Vision and Pattern Recognition}, pages 1334--1345, 2024.

\bibitem[Kim et~al.(2022)Kim, Kim, and Choi]{flame}
Jihoon Kim, Jiseob Kim, and Sungjoon Choi.
\newblock Flame: Free-form language-based motion synthesis \& editing.
\newblock \emph{arXiv preprint arXiv:2209.00349}, 2022.

\bibitem[Kingma and Welling(2013)]{vae}
Diederik~P Kingma and Max Welling.
\newblock Auto-encoding variational bayes.
\newblock \emph{arXiv preprint arXiv:1312.6114}, 2013.

\bibitem[Kutta(1901)]{Kutta}
W. Kutta.
\newblock Beitrag zur n\"aherungsweisen {I}ntegration totaler {D}ifferentialgleichungen.
\newblock \emph{Zeit. Math. Phys.}, 1901.

\bibitem[Le et~al.(2023)Le, Vyas, Shi, Karrer, Sari, Moritz, Williamson, Manohar, Adi, Mahadeokar, et~al.]{le2023voicebox}
Matthew Le, Apoorv Vyas, Bowen Shi, Brian Karrer, Leda Sari, Rashel Moritz, Mary Williamson, Vimal Manohar, Yossi Adi, Jay Mahadeokar, et~al.
\newblock Voicebox: Text-guided multilingual universal speech generation at scale.
\newblock In \emph{arXiv}, 2023.

\bibitem[Li et~al.(2022)Li, Zhao, Zhelun, and Sheng]{li2022danceformer}
Buyu Li, Yongchi Zhao, Shi Zhelun, and Lu Sheng.
\newblock Danceformer: Music conditioned 3d dance generation with parametric motion transformer.
\newblock \emph{AAAI}, 36\penalty0 (2):\penalty0 1272--1279, 2022.

\bibitem[Li et~al.(2024)Li, Zhang, Zhang, Zhang, Su, Guo, Liu, Liu, and Li]{li2024interdance}
Ronghui Li, Youliang Zhang, Yachao Zhang, Yuxiang Zhang, Mingyang Su, Jie Guo, Ziwei Liu, Yebin Liu, and Xiu Li.
\newblock Interdance: Reactive 3d dance generation with realistic duet interactions.
\newblock \emph{arXiv preprint arXiv:2412.16982}, 2024.

\bibitem[Liang et~al.(2023)Liang, Zhang, Li, Yu, and Xu]{intergen}
Han Liang, Wenqian Zhang, Wenxuan Li, Jingyi Yu, and Lan Xu.
\newblock Intergen: Diffusion-based multi-human motion generation under complex interactions.
\newblock \emph{arXiv preprint arXiv:2304.05684}, 2023.

\bibitem[Lipman et~al.(2023)Lipman, Chen, Ben-Hamu, Nickel, and Le]{lipman2022flow}
Yaron Lipman, Ricky~TQ Chen, Heli Ben-Hamu, Maximilian Nickel, and Matt Le.
\newblock Flow matching for generative modeling.
\newblock In \emph{ICLR}, 2023.

\bibitem[Lipman et~al.(2024)Lipman, Havasi, Holderrieth, Shaul, Le, Karrer, Chen, Lopez-Paz, Ben-Hamu, and Gat]{lipman2024flow}
Yaron Lipman, Marton Havasi, Peter Holderrieth, Neta Shaul, Matt Le, Brian Karrer, Ricky~TQ Chen, David Lopez-Paz, Heli Ben-Hamu, and Itai Gat.
\newblock Flow matching guide and code.
\newblock \emph{arXiv preprint arXiv:2412.06264}, 2024.

\bibitem[Liu et~al.(2019)Liu, Shahroudy, Perez, Wang, Duan, and Kot]{nturgbd120}
Jun Liu, Amir Shahroudy, Mauricio Perez, Gang Wang, Ling-Yu Duan, and Alex~C Kot.
\newblock Ntu rgb+ d 120: A large-scale benchmark for 3d human activity understanding.
\newblock \emph{T-PAMI}, 42\penalty0 (10):\penalty0 2684--2701, 2019.

\bibitem[Liu and Yi(2024)]{liu2024geneoh}
Xueyi Liu and Li Yi.
\newblock Geneoh diffusion: Towards generalizable hand-object interaction denoising via denoising diffusion.
\newblock \emph{arXiv preprint arXiv:2402.14810}, 2024.

\bibitem[Liu et~al.(2023{\natexlab{a}})Liu, Gong, and Liu]{rectifiedflow_iclr23}
Xingchao Liu, Chengyue Gong, and Qiang Liu.
\newblock Flow straight and fast: Learning to generate and transfer data with rectified flow.
\newblock In \emph{ICLR}, 2023{\natexlab{a}}.

\bibitem[Liu et~al.(2023{\natexlab{b}})Liu, Chen, and Yi]{liu2023interactive}
Yunze Liu, Changxi Chen, and Li Yi.
\newblock Interactive humanoid: Online full-body motion reaction synthesis with social affordance canonicalization and forecasting.
\newblock \emph{arXiv preprint arXiv:2312.08983}, 2023{\natexlab{b}}.

\bibitem[Liu et~al.(2024)Liu, Chen, Ding, and Yi]{liu2024physreaction}
Yunze Liu, Changxi Chen, Chenjing Ding, and Li Yi.
\newblock Physreaction: Physically plausible real-time humanoid reaction synthesis via forward dynamics guided 4d imitation.
\newblock In \emph{Proceedings of the 32nd ACM International Conference on Multimedia}, pages 3771--3780, 2024.

\bibitem[Martin et~al.(2024)Martin, Gagneux, Hagemann, and Steidl]{martin2024pnp}
S{\'e}gol{\`e}ne Martin, Anne Gagneux, Paul Hagemann, and Gabriele Steidl.
\newblock Pnp-flow: Plug-and-play image restoration with flow matching.
\newblock \emph{arXiv preprint arXiv:2410.02423}, 2024.

\bibitem[Neklyudov et~al.(2023)Neklyudov, Brekelmans, Severo, and Makhzani]{neklyudov2023action}
Kirill Neklyudov, Rob Brekelmans, Daniel Severo, and Alireza Makhzani.
\newblock Action matching: Learning stochastic dynamics from samples.
\newblock 2023.

\bibitem[Pavlakos et~al.(2019)Pavlakos, Choutas, Ghorbani, Bolkart, Osman, Tzionas, and Black]{smplx}
Georgios Pavlakos, Vasileios Choutas, Nima Ghorbani, Timo Bolkart, Ahmed~AA Osman, Dimitrios Tzionas, and Michael~J Black.
\newblock Expressive body capture: 3d hands, face, and body from a single image.
\newblock In \emph{CVPR}, pages 10975--10985, 2019.

\bibitem[Petrovich et~al.(2021)Petrovich, Black, and Varol]{petrovich2021action}
Mathis Petrovich, Michael~J Black, and G{\"u}l Varol.
\newblock Action-conditioned 3d human motion synthesis with transformer vae.
\newblock In \emph{ICCV}, 2021.

\bibitem[Raab et~al.(2022)Raab, Leibovitch, Li, Aberman, Sorkine-Hornung, and Cohen-Or]{modi}
Sigal Raab, Inbal Leibovitch, Peizhuo Li, Kfir Aberman, Olga Sorkine-Hornung, and Daniel Cohen-Or.
\newblock Modi: Unconditional motion synthesis from diverse data.
\newblock \emph{arXiv preprint arXiv:2206.08010}, 2022.

\bibitem[Ramesh et~al.(2022)Ramesh, Dhariwal, Nichol, Chu, and Chen]{ramesh2022hierarchical}
Aditya Ramesh, Prafulla Dhariwal, Alex Nichol, Casey Chu, and Mark Chen.
\newblock Hierarchical text-conditional image generation with clip latents.
\newblock \emph{arXiv preprint arXiv:2204.06125}, 2022.

\bibitem[Reitsma and Pollard(2003)]{reitsma2003perceptual}
Paul~SA Reitsma and Nancy~S Pollard.
\newblock Perceptual metrics for character animation: sensitivity to errors in ballistic motion.
\newblock In \emph{ACM SIGGRAPH 2003 Papers}, pages 537--542. 2003.

\bibitem[Rezende and Mohamed(2015)]{rezende2015variational}
Danilo Rezende and Shakir Mohamed.
\newblock Variational inference with normalizing flows.
\newblock In \emph{International conference on machine learning}, pages 1530--1538. PMLR, 2015.

\bibitem[Runge(1895)]{Runge}
Carl Runge.
\newblock {\"U}ber die numerische aufl{\"o}sung von differentialgleichungen.
\newblock \emph{Mathematische Annalen}, 1895.

\bibitem[Saharia et~al.(2022)Saharia, Chan, Saxena, Li, Whang, Denton, Ghasemipour, Gontijo~Lopes, Karagol~Ayan, Salimans, et~al.]{saharia2022photorealistic}
Chitwan Saharia, William Chan, Saurabh Saxena, Lala Li, Jay Whang, Emily~L Denton, Kamyar Ghasemipour, Raphael Gontijo~Lopes, Burcu Karagol~Ayan, Tim Salimans, et~al.
\newblock Photorealistic text-to-image diffusion models with deep language understanding.
\newblock \emph{Advances in neural information processing systems}, 35:\penalty0 36479--36494, 2022.

\bibitem[Siyao et~al.(2024)Siyao, Gu, Yang, Lin, Liu, Ding, Yang, and Loy]{siyao2024duolando}
Li Siyao, Tianpei Gu, Zhitao Yang, Zhengyu Lin, Ziwei Liu, Henghui Ding, Lei Yang, and Chen~Change Loy.
\newblock Duolando: Follower gpt with off-policy reinforcement learning for dance accompaniment.
\newblock \emph{arXiv preprint arXiv:2403.18811}, 2024.

\bibitem[Starke et~al.(2020)Starke, Zhao, Komura, and Zaman]{starke2020local}
Sebastian Starke, Yiwei Zhao, Taku Komura, and Kazi Zaman.
\newblock Local motion phases for learning multi-contact character movements.
\newblock \emph{TOG}, 39\penalty0 (4):\penalty0 54--1, 2020.

\bibitem[Tan et~al.()Tan, Li, Jin, Huang, Wang, and Song]{tanthink}
Wenhui Tan, Boyuan Li, Chuhao Jin, Wenbing Huang, Xiting Wang, and Ruihua Song.
\newblock Think then react: Towards unconstrained action-to-reaction motion generation.
\newblock In \emph{The Thirteenth International Conference on Learning Representations}.

\bibitem[Tanaka and Fujiwara(2023)]{role_aware}
Mikihiro Tanaka and Kent Fujiwara.
\newblock Role-aware interaction generation from textual description.
\newblock In \emph{Proceedings of the IEEE/CVF International Conference on Computer Vision}, pages 15999--16009, 2023.

\bibitem[Tang et~al.(2024)Tang, Wang, Ji, Xu, Yu, and Shi]{tang2024unified}
Jiangnan Tang, Jingya Wang, Kaiyang Ji, Lan Xu, Jingyi Yu, and Ye Shi.
\newblock A unified diffusion framework for scene-aware human motion estimation from sparse signals.
\newblock In \emph{Proceedings of the IEEE/CVF Conference on Computer Vision and Pattern Recognition}, pages 21251--21262, 2024.

\bibitem[Tevet et~al.(2023)Tevet, Raab, Gordon, Shafir, Cohen-or, and Bermano]{tevet2022human_mdm}
Guy Tevet, Sigal Raab, Brian Gordon, Yoni Shafir, Daniel Cohen-or, and Amit~Haim Bermano.
\newblock Human motion diffusion model.
\newblock In \emph{ICLR}, 2023.

\bibitem[Tian et~al.(2024)Tian, Yang, Ji, Ma, Xu, Yu, Shi, and Wang]{tian2024gaze}
Jie Tian, Lingxiao Yang, Ran Ji, Yuexin Ma, Lan Xu, Jingyi Yu, Ye Shi, and Jingya Wang.
\newblock Gaze-guided hand-object interaction synthesis: Benchmark and method.
\newblock \emph{arXiv e-prints}, pages arXiv--2403, 2024.

\bibitem[Trivedi et~al.(2021)Trivedi, Thatipelli, and Sarvadevabhatla]{ntu-x}
Neel Trivedi, Anirudh Thatipelli, and Ravi~Kiran Sarvadevabhatla.
\newblock Ntu-x: an enhanced large-scale dataset for improving pose-based recognition of subtle human actions.
\newblock In \emph{Proceedings of the Twelfth Indian Conference on Computer Vision, Graphics and Image Processing}, pages 1--9, 2021.

\bibitem[Vaswani et~al.(2017)Vaswani, Shazeer, Parmar, Uszkoreit, Jones, Gomez, Kaiser, and Polosukhin]{vaswani2017attention}
Ashish Vaswani, Noam Shazeer, Niki Parmar, Jakob Uszkoreit, Llion Jones, Aidan~N Gomez, {\L}ukasz Kaiser, and Illia Polosukhin.
\newblock Attention is all you need.
\newblock \emph{NeurIPS}, 2017.

\bibitem[Wang et~al.(2023{\natexlab{a}})Wang, Hunt, and Zhou]{wangdiffusion}
Zhendong Wang, Jonathan~J Hunt, and Mingyuan Zhou.
\newblock Diffusion policies as an expressive policy class for offline reinforcement learning.
\newblock In \emph{The Eleventh International Conference on Learning Representations}, 2023{\natexlab{a}}.

\bibitem[Wang et~al.(2023{\natexlab{b}})Wang, Wang, Lin, and Dai]{wang2023intercontrol}
Zhenzhi Wang, Jingbo Wang, Dahua Lin, and Bo Dai.
\newblock Intercontrol: Generate human motion interactions by controlling every joint.
\newblock \emph{CoRR}, 2023{\natexlab{b}}.

\bibitem[Wu et~al.(2023)Wu, Wang, Gong, Liu, Xiong, Ranjan, Krishnamoorthi, Chandra, and Liu]{wu2023fast}
Lemeng Wu, Dilin Wang, Chengyue Gong, Xingchao Liu, Yunyang Xiong, Rakesh Ranjan, Raghuraman Krishnamoorthi, Vikas Chandra, and Qiang Liu.
\newblock Fast point cloud generation with straight flows.
\newblock In \emph{CVPR}, 2023.

\bibitem[Wu et~al.(2024)Wu, Shi, Huang, Yu, Xu, and Wang]{wu2024thor}
Qianyang Wu, Ye Shi, Xiaoshui Huang, Jingyi Yu, Lan Xu, and Jingya Wang.
\newblock Thor: Text to human-object interaction diffusion via relation intervention.
\newblock \emph{arXiv preprint arXiv:2403.11208}, 2024.

\bibitem[Xu et~al.(2022)Xu, Song, Wang, Su, Fang, Ding, Gan, Yan, Jin, Yang, Zeng, and Wu]{actformer}
Liang Xu, Ziyang Song, Dongliang Wang, Jing Su, Zhicheng Fang, Chenjing Ding, Weihao Gan, Yichao Yan, Xin Jin, Xiaokang Yang, Wenjun Zeng, and Wei Wu.
\newblock Actformer: A gan-based transformer towards general action-conditioned 3d human motion generation.
\newblock \emph{arXiv e-prints}, pages arXiv--2203, 2022.

\bibitem[Xu et~al.(2023)Xu, Lv, Yan, Jin, Wu, Xu, Liu, Zhou, Rao, Sheng, Liu, Zeng, and Yang]{xu2023inter}
Liang Xu, Xintao Lv, Yichao Yan, Xin Jin, Shuwen Wu, Congsheng Xu, Yifan Liu, Yizhou Zhou, Fengyun Rao, Xingdong Sheng, Yunhui Liu, Wenjun Zeng, and Xiaokang Yang.
\newblock Inter-x: Towards versatile human-human interaction analysis.
\newblock \emph{arXiv preprint arXiv:2312.16051}, 2023.

\bibitem[Xu et~al.(2024)Xu, Zhou, Yan, Jin, Zhu, Rao, Yang, and Zeng]{xu2024regennet}
Liang Xu, Yizhou Zhou, Yichao Yan, Xin Jin, Wenhan Zhu, Fengyun Rao, Xiaokang Yang, and Wenjun Zeng.
\newblock Regennet: Towards human action-reaction synthesis.
\newblock In \emph{Proceedings of the IEEE/CVF Conference on Computer Vision and Pattern Recognition}, pages 1759--1769, 2024.

\bibitem[Yan et~al.(2018)Yan, Xiong, and Lin]{stgcn}
Sijie Yan, Yuanjun Xiong, and Dahua Lin.
\newblock Spatial temporal graph convolutional networks for skeleton-based action recognition.
\newblock In \emph{{AAAI}}, pages 7444--7452. {AAAI} Press, 2018.

\bibitem[Yan et~al.(2019)Yan, Li, Xiong, Yan, and Lin]{csgn}
Sijie Yan, Zhizhong Li, Yuanjun Xiong, Huahan Yan, and Dahua Lin.
\newblock Convolutional sequence generation for skeleton-based action synthesis.
\newblock In \emph{{ICCV}}, pages 4393--4401. {IEEE}, 2019.

\bibitem[Yang et~al.(2024)Yang, Ding, Cai, Yu, Wang, and Shi]{yang2024guidance}
Lingxiao Yang, Shutong Ding, Yifan Cai, Jingyi Yu, Jingya Wang, and Ye Shi.
\newblock Guidance with spherical gaussian constraint for conditional diffusion.
\newblock In \emph{Proceedings of the 41st International Conference on Machine Learning}, pages 56071--56095, 2024.

\bibitem[Yuan et~al.(2023)Yuan, Song, Iqbal, Vahdat, and Kautz]{yuan2023physdiff}
Ye Yuan, Jiaming Song, Umar Iqbal, Arash Vahdat, and Jan Kautz.
\newblock Physdiff: Physics-guided human motion diffusion model.
\newblock In \emph{Proceedings of the IEEE/CVF international conference on computer vision}, pages 16010--16021, 2023.

\bibitem[Zhang et~al.(2023{\natexlab{a}})Zhang, Zhang, Cun, Huang, Zhang, Zhao, Lu, and Shen]{zhang2023generating_t2mgpt}
Jianrong Zhang, Yangsong Zhang, Xiaodong Cun, Shaoli Huang, Yong Zhang, Hongwei Zhao, Hongtao Lu, and Xi Shen.
\newblock T2m-gpt: Generating human motion from textual descriptions with discrete representations.
\newblock In \emph{CVPR}, 2023{\natexlab{a}}.

\bibitem[Zhang et~al.(2022)Zhang, Cai, Pan, Hong, Guo, Yang, and Liu]{motiondiffuse}
Mingyuan Zhang, Zhongang Cai, Liang Pan, Fangzhou Hong, Xinying Guo, Lei Yang, and Ziwei Liu.
\newblock Motiondiffuse: Text-driven human motion generation with diffusion model.
\newblock \emph{arXiv preprint arXiv:2208.15001}, 2022.

\bibitem[Zhang et~al.(2023{\natexlab{b}})Zhang, Guo, Pan, Cai, Hong, Li, Yang, and Liu]{zhang2023remodiffuse}
Mingyuan Zhang, Xinying Guo, Liang Pan, Zhongang Cai, Fangzhou Hong, Huirong Li, Lei Yang, and Ziwei Liu.
\newblock Remodiffuse: Retrieval-augmented motion diffusion model.
\newblock In \emph{ICCV}, 2023{\natexlab{b}}.

\bibitem[Zheng et~al.(2023)Zheng, Le, Shaul, Lipman, Grover, and Chen]{zheng2023guided}
Qinqing Zheng, Matt Le, Neta Shaul, Yaron Lipman, Aditya Grover, and Ricky~TQ Chen.
\newblock Guided flows for generative modeling and decision making.
\newblock \emph{arXiv preprint arXiv:2311.13443}, 2023.

\bibitem[Zhou et~al.(2022)Zhou, Bhatnagar, Lenssen, and Pons-Moll]{zhou2022toch}
Keyang Zhou, Bharat~Lal Bhatnagar, Jan~Eric Lenssen, and Gerard Pons-Moll.
\newblock Toch: Spatio-temporal object-to-hand correspondence for motion refinement.
\newblock In \emph{European Conference on Computer Vision}, pages 1--19. Springer, 2022.

\bibitem[Zhou et~al.(2019)Zhou, Barnes, Lu, Yang, and Li]{6d_rot}
Yi Zhou, Connelly Barnes, Jingwan Lu, Jimei Yang, and Hao Li.
\newblock On the continuity of rotation representations in neural networks.
\newblock In \emph{{CVPR}}, pages 5745--5753. Computer Vision Foundation / {IEEE}, 2019.

\bibitem[Zhou and Wang(2023)]{zhou2023ude}
Zixiang Zhou and Baoyuan Wang.
\newblock Ude: A unified driving engine for human motion generation.
\newblock In \emph{Proceedings of the IEEE/CVF conference on computer vision and pattern recognition}, pages 5632--5641, 2023.

\end{thebibliography}

\clearpage
\setcounter{page}{1}
\onecolumn
\begin{center}
{ \linespread{2.0} \selectfont
\textbf{\Large \methodname: Human Action-Reaction Flow Matching with\\ Physical Guidance} \\
}
\linespread{1.5} \Large Supplementary Materials 
\end{center}
\appendix
\setcounter{page}{1}

\counterwithin{table}{section}
\counterwithin{figure}{section}

\makeatletter

\startcontents[appendix]
\printcontents[appendix]{l}{1}{\section*{Appendix}\setcounter{tocdepth}{2}}

\clearpage


\section{Algorithm derivation}
\label{append:derivation}
We denote the deterministic functions: $\hat{\bx}_{1}=\mathbb{E}[\bx_1|\bx_t,c]$ as the $\bx_{1}$-prediction, $\bv_\theta(\bx_t,t,c)=u_{t}(\bx_t)$ as the $\bv$-prediction.
By defining the conditional probability path as a linear interpolation between $p_0$ and $p_1$, the intermediate process becomes: 
\begin{align}
\label{eq:inter}
\bx_t = t \bx_1 + [1 - (1 - \sigma_\text{min}) t] \bx_0,~~~~~~~~~
\end{align}
\noindent where $\sigma_{\text{min}}>0$ is a small amount of noise. Take the derivative of $t$ on both sides:
\begin{align}
\label{eq:marginal}
\frac{d\bx_t}{dt}=\bx_1 - (1 - \sigma_\text{min})\bx_0,~~~~~~~
\end{align}
In the marginal velocity formula, we obtain:
\begin{align}
\label{eq:1}
u_{t}(\bx_t)&= \mathbb{E}[\bx_1 - (1 - \sigma_\text{min})\bx_0|\bx_t,c] \nonumber \\
&= \mathbb{E}[\bx_1|\bx_t,c] - (1 - \sigma_\text{min})\mathbb{E}[\bx_0|\bx_t,c].~~~~
\end{align}
Substitute $\bx_0 = \frac{\bx_t - t \bx_1}{1 - (1 - \sigma_\text{min}) t}$ from Eq.~\ref{eq:inter} into the above equation:
\begin{align}
~~~~~~~u_{t}(\bx_t)&= \mathbb{E}[\bx_1|\bx_t,c] - (1 - \sigma_\text{min}) \frac{\bx_t - t~\mathbb{E}[\bx_1|\bx_t,c]}{1 - (1 - \sigma_\text{min}) t} \nonumber \\
&=\frac{\mathbb{E}[\bx_1|\bx_t,c] - (1 - \sigma_\text{min})\bx_t}{1 - (1 - \sigma_\text{min})t} \nonumber
\end{align}
where we have used the fact that $E [\bx_t|\bx_t ] = \bx_t$. 
According to $\hat{\bx}_{1}=\mathbb{E}[\bx_1|\bx_t,c]$ , $\bv_\theta(\bx_t,t,c)=u_{t}(\bx_t)$, we get the equivalent form of parameterization:
\begin{align}
\bv_\theta(\bx_t,t,c) &= \frac{\hat{\bx}_{1} - (1 - \sigma_\text{min})\bx_t}{1 - (1 - \sigma_\text{min})t},
\label{eq:transform}
\end{align}
Substitute~Eq.~\ref{eq:transform} into the following equation:
\begin{align}
~~~~~~~~~~~~\bx_{t'}&=\bx_{t}-(t-t')\bv_\theta(\bx_t,t,c) \nonumber\\
&=\bx_{t}-(t-t')\frac{\hat{\bx}_{1} - (1 - \sigma_\text{min})\bx_t}{1 - (1 - \sigma_\text{min})t} \nonumber\\
&=\frac{1-(1 - \sigma_\text{min})t'}{1 - (1 - \sigma_\text{min})t}\bx_{t} + \frac{t'-t}{1 - (1 - \sigma_\text{min})t}\hat{\bx}_{1}.
\label{eq:4}
\end{align}
Finally, let $t = t_{n}$ and $t' = t_{n+1}$, we can obtain the estimation of $\hat{\bx}_{1}$ from~Eq.~\ref{eq:transform}: 
\begin{align}
\label{v-prediction}
~~~~~~~~&\hat{\bx}_1 \leftarrow (1 - \sigma_\text{min})\bx_{t_{n}} + (1-(1 - \sigma_\text{min})t_{n})~\bv_{\theta}(\bx_{t_{n}}, t_{n},\bc), 
\end{align}
and our sampling formulation based on $\bx_1$-prediction from~Eq.~\ref{eq:4}:
\begin{align}
\label{eq:append-x1}
~~~~~~~~&\bx'_{t_{n+1}} \leftarrow \frac{1-(1 - \sigma_\text{min})t_{n+1}}{1-(1 - \sigma_\text{min})t_{n}} \bx_{t_{n}} + \frac{t_{n+1}-t_{n}}{1-(1 - \sigma_\text{min})t_{n}} ~\hat{\bx}_1. 
\end{align}

\begin{algorithm}[h]
\small
\caption{
Sampling algorithm with vanilla guidance of physical constraints. ($\bv$-prediction)}
\label{alg:guidance_v}
\begin{algorithmic}[1]
\STATE{\textbf{Input}: $\mathcal{L}_\text{pene}$ the loss function ; $\bv$ and $\theta$ the vector field predictor with pretrained parameters}
\STATE{\textbf{Parameters}: $N$ the number of sampling steps; 
$\lambda_\text{pene}$ the scale factor to control the strength of guidance }
\STATE{Sample $\bx_0$ from the action distribution}
\FOR{$n=1,2,...,N-1$}
\STATE \comment{\small \# Estimate $\hat{\bx}_1$ (Eq. \ref{v-prediction})}
\STATE{$\hat{\bx}_1 \leftarrow (1 - \sigma_\text{min})\bx_{t_{n}} + (1-(1 - \sigma_\text{min})t_{n})~\bv_{\theta}(\bx_{t_{n}}, t_{n},\bc) $}
\STATE \comment{\small \# Flow maching $\bv$-prediction sampling (Eq. \ref{eq:sample})}
\STATE{$\bx'_{t_{n+1}} \leftarrow \bx_{t_{n}} + (t_{n+1}-t_{n}) ~\bv_{\theta}(\bx_{t_{n}}, t_{n},\bc)$}
\STATE \comment{\small \# Physical constraint guidance}
\STATE{$\bx_{t_{n+1}} \leftarrow \bx'_{t_{n+1}} + \lambda_\text{pene}\nabla_{\bx_{t_{n}}} \mathcal{L}_\text{pene}(\hat{\bx}_1)$}
\ENDFOR
\STATE{\textbf{Return}: The reaction motion after guidance $\bx_{1}=\bx_{t_N} $}
\end{algorithmic}
\end{algorithm}

\section{Experimental results on the InterHuman-AS dataset and offline settings}
\label{sec:interhuman}
For the text-conditioned setting, we adopt T2M~\cite{guo2022generating}, MDM~\cite{tevet2022human_mdm}, MDM-GRU~\cite{tevet2022human_mdm}, RAIG~\cite{role_aware} and InterGen~\cite{intergen} as baselines. Tab.~\ref{tab:cmdm_interhuman} shows our method also yields the best results compared to baselines. To demonstrate the universality of our ARFlow and physical guidance methods, we also conducted offline setting experiments in~Tab.\ref{tab:cmdm_offline} and Tab.\ref{tab:ablations_offline}.

\begin{table}[h]
\centering
\begin{center}
\resizebox{0.9\textwidth}{!}{
\begin{tabular}{ l c c c c c}
  \toprule
  \multirow{2}{1cm}{Methods}  & \multirow{2}{2.cm}{\centering R Precision (Top 3)$\uparrow$} & \multirow{2}{1.5cm}{\centering FID $\downarrow$} & \multirow{2}{2.5cm}{\centering MM Dist$\downarrow$}  & \multirow{2}{2cm}{\centering Diversity$\rightarrow $} & \multirow{2}{*}{\centering MModality $\uparrow$}\\
  \\
  \midrule
    Real & $0.722^{\pm0.004}$ & $0.002^{\pm0.0002}$ & $3.503^{\pm0.011}$ & $5.390^{\pm0.058}$ & - \\
    \midrule
    T2M~\cite{guo2022generating} & $0.224^{\pm0.003}$ & $32.482^{\pm0.0975}$ & $7.299^{\pm0.016}$ & $4.350^{\pm0.073}$ & $0.719^{\pm0.041}$ \\
    MDM~\cite{tevet2022human_mdm} & $0.370^{\pm0.006}$ & $3.397^{\pm0.0352}$ & $8.640^{\pm0.065}$ & $4.780^{\pm0.117}$ & $2.288^{\pm0.039}$\\
    MDM-GRU~\cite{tevet2022human_mdm} & $0.328^{\pm0.012}$ & $6.397^{\pm0.2140}$ & $8.884^{\pm0.040}$ & $4.851^{\pm0.081}$ & $2.076^{\pm0.040}$\\
    RAIG~\cite{role_aware} & $0.363^{\pm0.008}$ & $2.915^{\pm0.0292}$ & $7.294^{\pm0.027}$ & $4.736^{\pm0.099}$ & $2.203^{\pm0.049}$\\
    InterGen~\cite{intergen} & $0.374^{\pm0.005}$ & $13.237^{\pm0.0352}$ & $10.929^{\pm0.026}$ & $4.376^{\pm0.042}$ & $\mathbf{2.793^{\pm0.014}}$ \\
    ReGenNet~\cite{xu2024regennet} & $0.407^{\pm0.003}$ & $2.265^{\pm0.0969}$ & $6.860^{\pm0.0040}$ & $5.214^{\pm0.139}$ & $2.391^{\pm0.023}$\\
    \midrule
    ARFlow & $\mathbf{0.434^{\pm0.003}}$ & $\mathbf{1.637^{\pm0.0413}}$ & $\mathbf{3.949^{\pm0.0042}}$ & $\mathbf{5.259^{\pm0.117}}$ & $2.502^{\pm0.021}$\\
  \bottomrule
\end{tabular}
}
\end{center}
\caption{\textbf{Comparison to state-of-the-arts} on the \textit{online, unconstrained} setting for human action-reaction synthesis on the InterHuman-AS dataset.  $\rightarrow$ denotes that the result closer to Real is better, and $\pm$ represents 95\% confidence interval. We highlight the best result in \textbf{Bold}.}
\label{tab:cmdm_interhuman}
\end{table}

\begin{table*}[hbtp]
  \vspace{-1mm}
  \begin{center}
  \resizebox{0.9\textwidth}{!}{
  \begin{tabular}{l c c c c }
  \toprule
   Method & FID$\downarrow$ & Acc.$\uparrow$ & Div.$\rightarrow$ & Multimod.$\rightarrow$ \\
  \midrule
  Real & $0.09^{\pm0.00}$ & $0.867^{\pm0.0002}$ & $13.06^{\pm0.09}$ & $25.03^{\pm0.23}$ \\
  \midrule
  cVAE~\cite{vae} & $74.73^{\pm4.86}$ & $0.760^{\pm0.0002}$ & $11.14^{\pm0.04}$ & $18.40^{\pm0.26}$ \\
  AGRoL~\cite{agrol} & $16.55^{\pm1.41}$ & $0.716^{\pm0.0002}$ & $13.84^{\pm0.10}$ & $21.73^{\pm0.20}$ \\
  MDM~\cite{tevet2022human_mdm} & $ 7.49^{\pm0.62}$ & $\mathbf{0.775^{\pm0.0003}}$ & $\underline{13.67^{\pm0.18}}$ & $24.14^{\pm0.29}$ \\
  MDM-GRU~\cite{tevet2022human_mdm} & $24.25^{\pm1.39}$ & $0.720^{\pm0.0002}$ & $\mathbf{13.43^{\pm0.09}}$ & $22.24^{\pm0.29}$ \\
  ReGenNet~\cite{xu2024regennet} &  $\underline{6.19^{\pm0.33}}$ & $ 0.772^{\pm0.0003}$ & $14.03^{\pm0.09}$ & $\underline{25.21^{\pm0.34}}$ \\
  \textbf{\methodname} & $\mathbf{5.00^{\pm0.17}}$ & $\underline{0.772^{\pm0.0002}}$ & $13.84^{\pm0.09}$ & $\mathbf{25.10^{\pm0.17}}$ \\
  \bottomrule
  \end{tabular}
  }
  \caption[caption]{\textbf{Results} on the \textbf{offline}\textit{, unconstrained} setting on NTU120-AS. We highlight the best result in \textbf{Bold} and the second best in \underline{underline}.}
  \label{tab:cmdm_offline}
  \end{center}
  \end{table*}

\begin{table*}[htbp]
  \vspace{-1mm}
  \begin{center}
  \resizebox{0.9\textwidth}{!}{
  \begin{tabular}{l l c c c c c }
  \toprule
  \multirow{2}{*}{Class} & \multirow{2}{*}{Settings}  & \multirow{2}{*}{FID$\downarrow$} & \multirow{2}{*}{Acc.$\uparrow$} & \multirow{2}{*}{Div.$\rightarrow$} & \multirow{2}{*}{Multimod.$\rightarrow$}  & \multirow{2}{*}{Latency(ms)}\\
  & \\
  \midrule
  & Real & $0.085^{\pm0.0003}$ & $0.867^{\pm0.0002}$ & $13.063^{\pm0.0908}$ & $25.032^{\pm0.2332}$ & - \\
  \midrule
  \multirow{2}{*}{Prediction} & 1) $x_1$  & $\mathbf{5.003^{\pm0.1654}}$ & $\mathbf{0.762^{\pm0.0002}}$ & $13.844^{\pm0.0905}$ & $\mathbf{25.104^{\pm0.1704}}$ & - \\
  &2) $v$ & $7.585^{\pm0.1562}$ & $0.757^{\pm0.0002}$ & $13.775^{\pm0.0982}$ & $24.200^{\pm0.1355}$ & - \\
  \midrule
  \multirow{1}{*}{Guidance} & w. $\mathcal{L}_\text{pene}$ & $5.048^{\pm0.1167}$ & $0.750^{\pm0.0002}$ & $13.838^{\pm0.0893}$ & $25.048^{\pm0.1595}$ & - \\
  \midrule
  \multirow{4}{*}{Timesteps} & 2 & $7.936^{\pm0.1581}$ & $0.759^{\pm0.0002}$ & $14.538^{\pm0.1016}$ & $25.904^{\pm0.1754}$ & $0.023$\\
  & 5 & $\mathbf{5.003^{\pm0.1654}}$ & $\mathbf{0.762^{\pm0.0002}}$ & $13.844^{\pm0.0905}$ & $\mathbf{25.104^{\pm0.1704}}$ & $0.053$\\
  & 10 & $5.506^{\pm0.1657}$ & $0.744^{\pm0.0002}$ & $13.870^{\pm0.0942}$ & $24.732^{\pm0.1533}$ & $0.110$\\
  & 100 & $5.836^{\pm0.3763}$ & $0.748^{\pm0.0002}$ & $13.635^{\pm0.0948}$ & $24.058^{\pm0.1371}$ & $1.132$\\
  \midrule
  Best &\textbf{\methodname} & $\mathbf{5.003^{\pm0.1654}}$ & $\mathbf{0.762^{\pm0.0002}}$ & $13.844^{\pm0.0905}$ & $\mathbf{25.104^{\pm0.1704}}$ & $0.053$ \\
  \bottomrule
  \end{tabular}
  }
  \end{center}
  \caption[caption]{\textbf{Ablation studies} on the \textbf{offline}\textit{, unconstrained} setting on the NTU120-AS dataset. \textbf{Bold} indicates the best result in our method.}
  \label{tab:ablations_offline}
  \end{table*}

\section{Influence of sampling randomness}
As depicted in Tab.~\ref{tab:randomness}, although stochastic sampling increases the diversity of generated reaction motions, it sometimes has some impact on the quality of the sample due to its stochastic nature.

\begin{table*}[t]
  \vspace{-1mm}
  \begin{center}
  \resizebox{0.9\textwidth}{!}{
  \begin{tabular}{l l c c c c }
  \toprule
  \multirow{2}{*}{Method} & \multirow{2}{*}{Settings}  & \multirow{2}{*}{FID$\downarrow$} & \multirow{2}{*}{Acc.$\uparrow$} & \multirow{2}{*}{Div.$\rightarrow$} & \multirow{2}{*}{Multimod.$\rightarrow$} \\
  & \\
  \midrule
    & Real & $0.085^{\pm0.0003}$ & $0.867^{\pm0.0002}$ & $13.063^{\pm0.0908}$ & $25.032^{\pm0.2332}$ \\
  \midrule
   \multirow{3}{*}{Randomness $\beta$ } & 0.05 & $13.821^{\pm0.2895}$ & $0.709^{\pm0.0003}$ & $14.002^{\pm0.1055}$ & $24.269^{\pm0.1363}$  \\
    & 0.02 & $8.060^{\pm0.1517}$ & $0.728^{\pm0.0002}$ & $13.928^{\pm0.1076}$ & $24.161^{\pm0.1512}$  \\
    & 0.01 & $7.671^{\pm0.1357}$ & $0.728^{\pm0.0002}$ & $13.895^{\pm0.1080}$ & $24.114^{\pm0.1486}$  \\
  \midrule
  \methodname & 0 & $7.894^{\pm0.1814}$ & $\mathbf{0.743^{\pm0.0002}}$ & $ \mathbf{13.599^{\pm0.1005}}$ & $24.105^{\pm0.1310}$ \\
  \bottomrule
  \end{tabular}
  }
  \end{center}
  \caption[caption]{\textbf{Randomness Influence studies} on the \textit{online}\textit{, unconstrained} setting on the NTU120-AS dataset. \textbf{Bold} indicates the best result in our method.}
  \label{tab:randomness}
  \end{table*}

\section{Parameter analysis of guidance strength and weight factor}
We conduct a parameter analysis of guidance strength in Tab.~\ref{table:strength}. The result of the experiments show that as the guiding strength increases, the degree of penetration  between actors and reactors decreases significantly, while FID increases slightly. This is because the ground truth itself has a certain degree of penetration. Thus, this task requires our new metrics and FID to collaborate in evaluating the quality of the generated results. When the guidance strength increases to a certain extent, the decrease in penetration degree is no longer significant. Therefore, we ultimately choose $\lambda_\text{pene}=2$. Our method achieves the lowest penetration level while maintaining the best generation quality.
As for the weight factor, the results show that the minimum value of FID does not occur at the endpoints, thus demonstrating the effectiveness of our weighting method.

\begin{table}[h] 
\caption{Parameter analysis of guidance strength and weight factor on the \textit{online}\textit{, unconstrained} setting on NTU120-AS. \textbf{Bold} indicates the best result.}
\vspace{0.5em}
\footnotesize
\centering
\begin{tabular}{lccccc}
\toprule
 Parameter& settings & FID $\downarrow$~ &  IF $\downarrow$  & IV $\downarrow$\\ 
 \midrule
 & Real & 0.09 & 21.96\% & 5.35 \\ 
 \midrule
\multirow{5}{*}{~~~~~$\lambda_\text{pene}$} & 0& 7.89  & 8.39\% &  3.26 & \\
& 1& 7.98  & 5.80\% & 1.15 & \\  
& \textbf{2}& \textbf{8.20}  & \textbf{3.54\%} & \textbf{0.68} & \\  
& 5& 8.49 & 1.22\% & 0.13 \\ 
& 10 & 9.41 & 0.78\% & 0.21 \\ 
\midrule
\multirow{7}{*}{~~~~~$w$} & 0& 8.37  & 2.71\% &  0.35 & \\
& 0.1& 8.30  & 2.78\% & 0.36 & \\  
& 0.3& 8.19  & 2.96\% & 0.37 & \\  
& 0.5& 8.11 & 3.12\% & 0.41 &\\ 
& \textbf{0.7} & \textbf{8.07} & \textbf{3.23\%} & \textbf{0.53} \\   
& 0.9& 8.08  & 3.32\% & 0.64 & \\  
& 1& 8.20 & 3.54\% & 0.68 \\ 
\bottomrule
\end{tabular}
\label{table:strength}
\vspace{-1em}
\end{table}

\section{Details of our framework}
\label{sec:arflow}
We present our Human Action-Reaction Flow Matching (\methodname) framework, illustrated in Fig.~\ref{fig:pipeline1}, which comprises a flow module and a Transformer decoder $G$. Given a paired action-reaction sequence and an optional signal $c$  (e.g. , an action label, dotted lines in Fig.~\ref{fig:pipeline1}), $<$$x_0^{1:H}$, $x_1^{1:H}$, $c$$>$, $x_1^{1:H}$ represents the reaction to generate. For a sampled timestep $t$, we linearly interpolate $x_1^{1:H}$ and $x_0^{1:H}$ as Eq.~\ref{eqn:flow} to produce the $x^{1:H}_t$. Then the \textit{$x^{1:H}_t$ } turns into the latent features through an FC layer to dimension \textit{d}. The timestep $t$ and the optional condition $c$ are separately projected to dimension $d$ using feed-forward networks and combined to form the token $z$.
The Transformer decoder~$G$, implemented with stacked 8 layers, prevents future information leakage through masked multi-head attention, enabling \textit{online} generation as in~\cite{xu2024regennet}. Decoder $G$ takes $z$ as input tokens and $x^{1:H}_t$ combined with a standard positional embedding as output tokens, along with a directional attention mask to ensure the model cannot access future actions at the current timestep. The decoder's output is projected back to produce the predicted clean body poses $\hat{x}_1^{1:H}$. Online reaction generation is achieved in an auto-regressive manner, following the approach of~\cite{xu2024regennet}.
The intention branch can be activated when the actor's intention is accessible to the reactor, or deactivated otherwise. The directional attention mask can be turned off for offline settings.

At the inference stage, we employ our physical constraint guidance method.
After training latent linear layers and Transformer decoder $G$, our ARFlow uses them for $x_1$-prediction based sampling. The sampling process is further guided by the gradient of $\mathcal{L}_{pene}$ to generate physically plausible reactions.

\section{Implementation details}
\label{appendix:details}
Our \methodname~model is trained with $T=1,000$ timesteps using a classifier-free approach~\cite{ho2021classifier}. The number of decoder layers is 8 and the latent dimension of the Transformer tokens is 512. The batch size is configured as 32 for NTU120-AS, InterHuman-AS and 16 for Chi3D-AS. The interaction loss weight is set to $\lambda_\text{inter}=1$.
Each model is trained for 500K steps on single NVIDIA 4090 GPU within 48 hours.
During inference, unless otherwise stated, we employ 5-timestep sampling for all the diffusion-based and our models in our experiments as~\cite{xu2024regennet} for a fair comparison. For the physical constraint guidance, we set the safe distance  $\zeta= 0.5$, $\lambda_\text{pene}=2$ and $w=0.7$ .

\begin{figure}[h]
  \centering
  \includegraphics[width=0.7\linewidth]{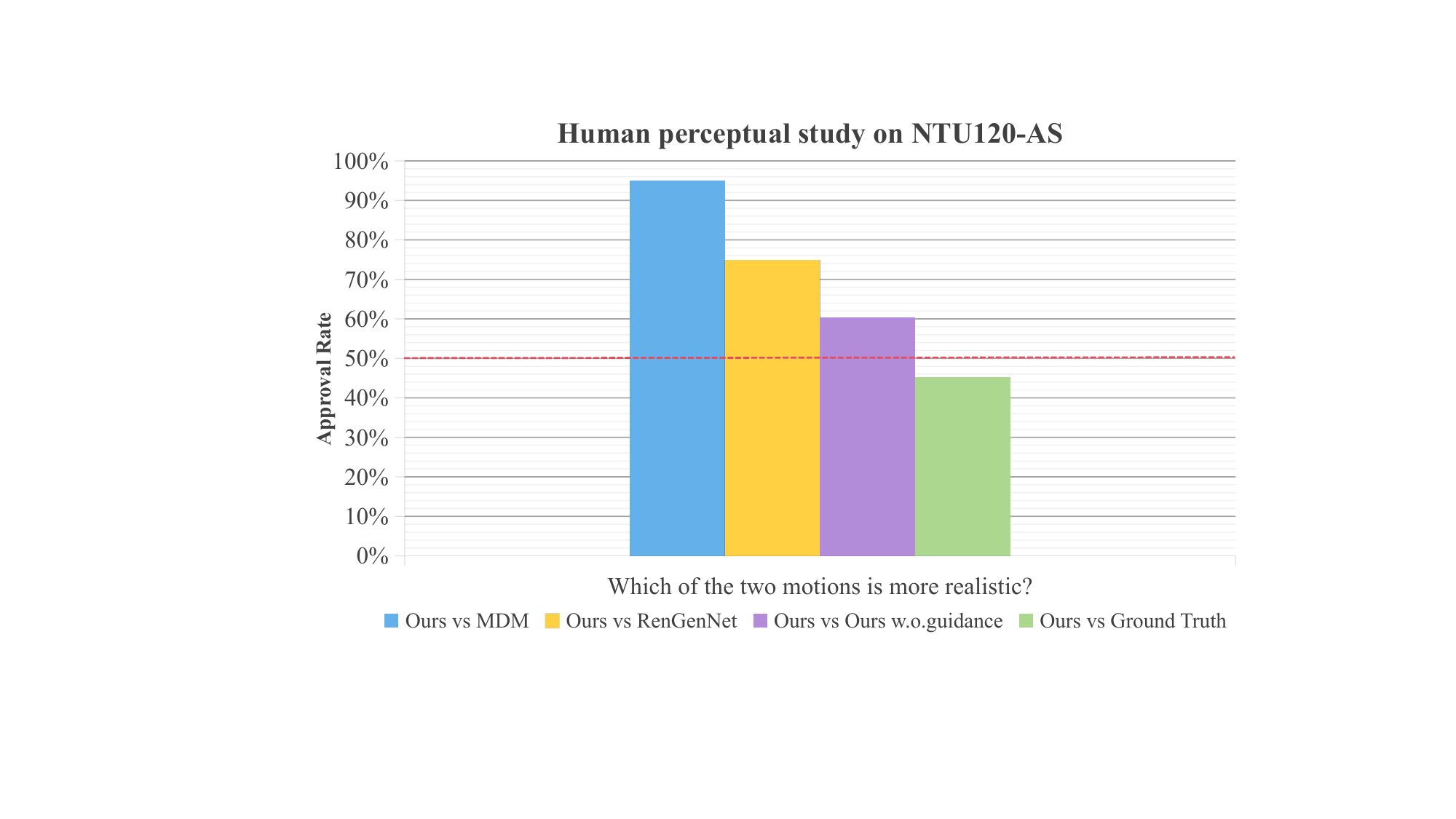}
  \vspace{-4mm}
  \caption{Human perceptual study results on NTU120-AS.}
  \label{fig:vis11}
  \vspace{-2mm}
\end{figure}

\begin{figure}[h]
  \centering
  \includegraphics[width=0.7\linewidth]{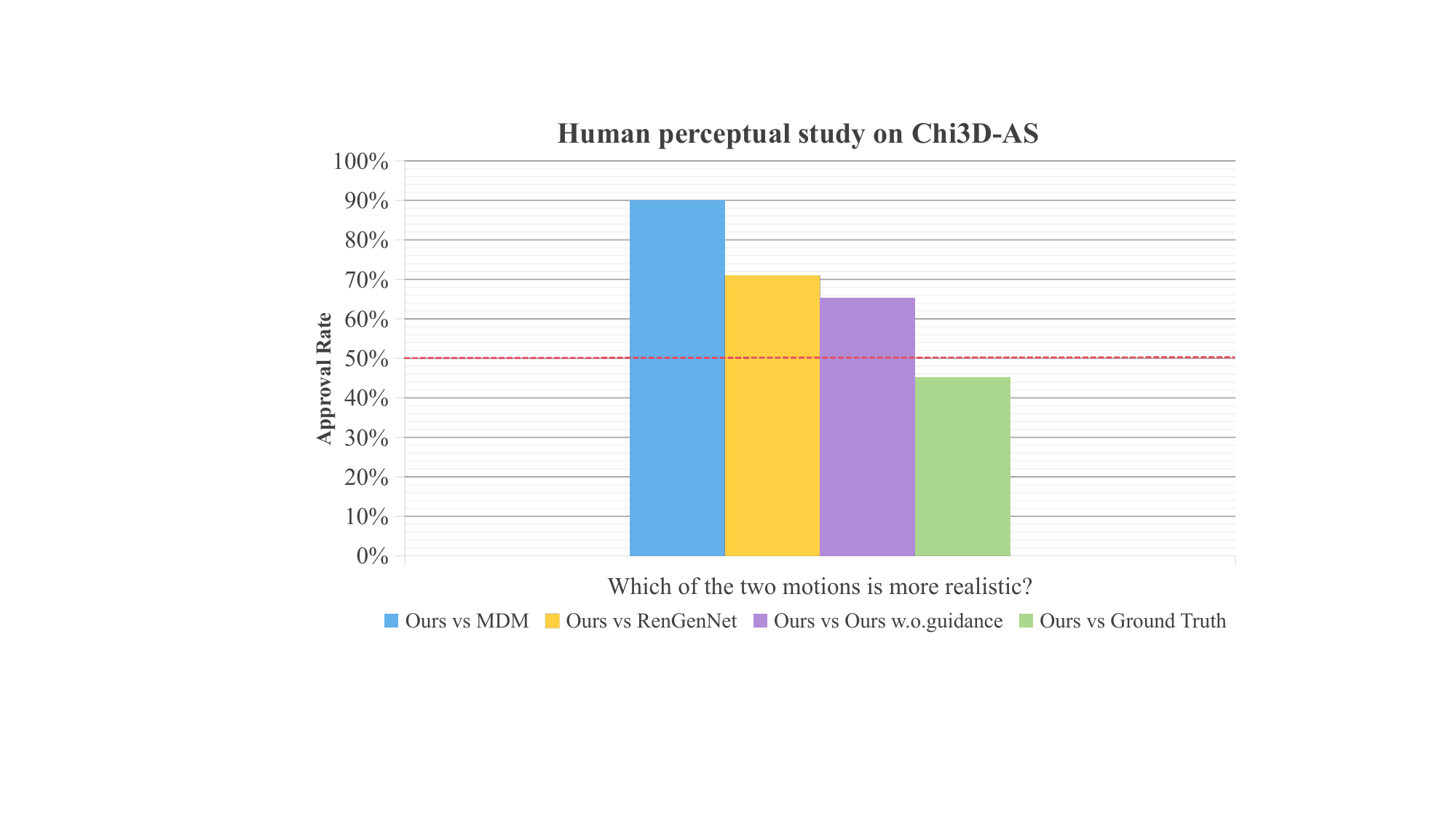}
  \vspace{-4mm}
  \caption{Human perceptual study results on Chi3D-AS.}
  \label{fig:vis22}
  \vspace{-2mm}
\end{figure}

\section{Details of the metric calculations} 
We follow the prior works in human action-reaction synthesis, ReGenNet~\cite{xu2024regennet} and MDM~\cite{tevet2022human_mdm} to calculate the Frechet Inception Distance(FID)~\cite{fid}, action recognition accuracy, diversity and multi-modality. For a fair comparison, we use the pre-trained action recognition model in~\cite{xu2024regennet}, which is a slightly modified version of ST-GCN~\cite{stgcn}. The model takes the 6D rotation representation of the SMPL-X parameters as input and outputs classification results of action-reaction pairs. We generate 1,000 reaction samples by sampling actor motions from test sets and evaluate each method 20 times using different random seeds to calculate the average with the 95\% confidence interval.

1)~Frechet Inception Distance (FID) \cite{fid} measures the similarity in feature space between predicted and ground-truth motion;
2)~Action Recognition Accuracy (Acc.) assesses how likely a generated motion can be successfully recognized. We adopt the pre-trained ST-GCN model to classify the generated results;
3) Diversity (Div.) evaluates feature diversity within generated motions.
Given the motion feature vectors of generated motions and real motions as \{$v_1, \cdots, v_{S_d}$\} and \{$v_1^\prime, \cdots, v_{S_d}^\prime$\}, the diversity is defined as $Diversity=\frac{1}{S_d}\sum_{i=1}^{S_d}||v_i-v_i^\prime||_2$. $S_d=200$ in our experiments.
4) Multi-modality (Multimod.) quantifies the ability to generate multiple different motions for the same action type. Given a collection of motions containing $C$ action types, for $c$-th action, we randomly sample two subsets of size $S_l$, and then extract the corresponding feature vectors as \{$v_{c,1}, \cdots, v_{c, S_l}$\} and \{$v_{c,1}^\prime, \cdots, v_{c,S_l}^\prime$\}, the multimodality is defined as $Multimod.=\frac{1}{C\times S_l}\sum_{c=1}^C\sum_{i=1}^{S_l}||v_{c,i}-v_{c,i}^\prime||_2$. $S_l=20$ in our experiments. 5) Intersection Volume (IV) measures the volume of human-human inter-penetration by voxelizing actor and reactor meshes and reporting the volume of voxels occupied by both.  
6) Intersection Frequency (IF) measures the frequency of inter-penetration. 
We generate 260 samples for evaluation.



\section{User Study}
We conducted a human perceptual study to investigate the
quality of the motions generated by our model. We invite 20
users to provide four comparisons. For each comparison, we
ask the users ``Which of the two motions is more realistic?”,
and each user is provided 10 sequences to evaluate.

The results are shown in Fig.~\ref{fig:vis11} and Fig.~\ref{fig:vis22}. Our
results were preferred over the other state-of-the-art and are even competitive with ground truth motions.

\begin{figure}[t]
  \centering
  \includegraphics[width=1\linewidth]{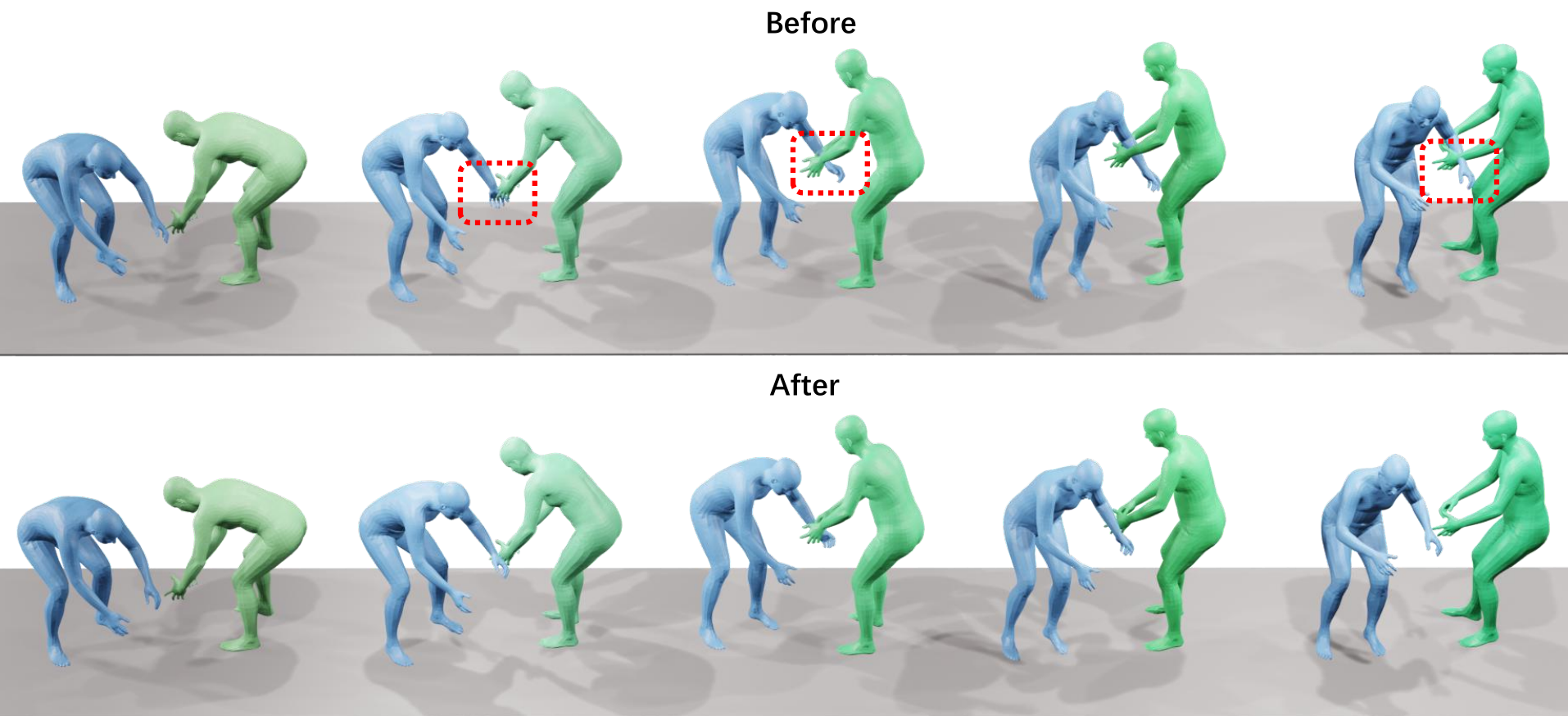}
  \vspace{-4mm}
  \caption{\textbf{Visualization comparison} of the effects before and after using physical constraint guidance. \textcolor{blue}{Blue} for actors and \textcolor{green}{Green} for reactors.}
  \label{fig:vis}
  \vspace{-2mm}
\end{figure}

\section{Extra qualitative results}
\begin{figure}[h]
  \centering
  \includegraphics[width=1\linewidth]{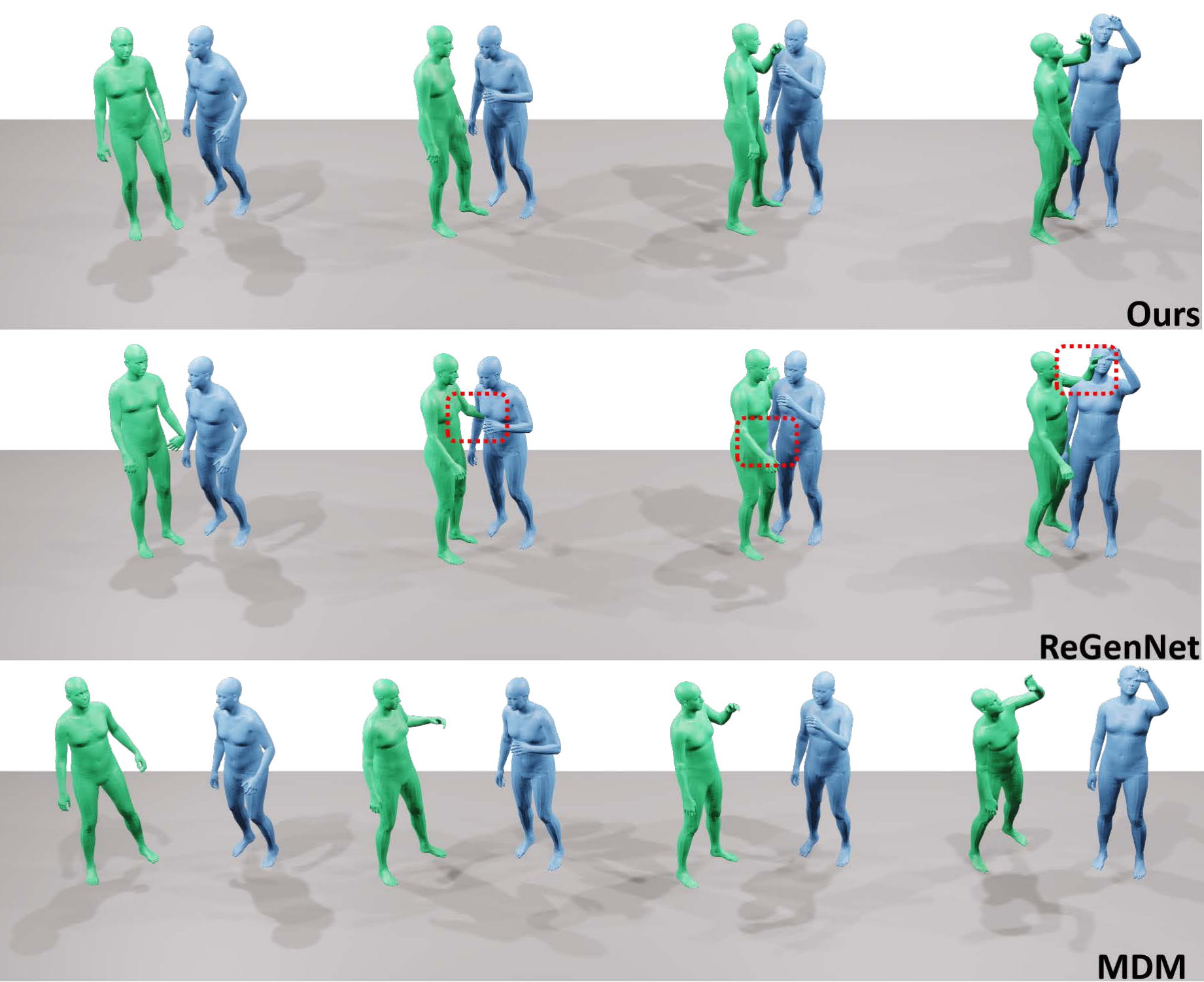}
  \caption{The extra qualitative experiment. \textcolor{blue}{Blue} for actors and \textcolor{green}{Green} for reactors.}
  \label{fig:extra}
\end{figure}

We show the generated motions of our method against
others in Fig.~\ref{fig:extra}. We highlight the
implausible motions in rectangle marks, it is clear that our
method learns the correct reactions and
avoids human-human inter-penetrations as much as possible.

\begin{figure}[h]
  \centering
  \includegraphics[width=1\linewidth]{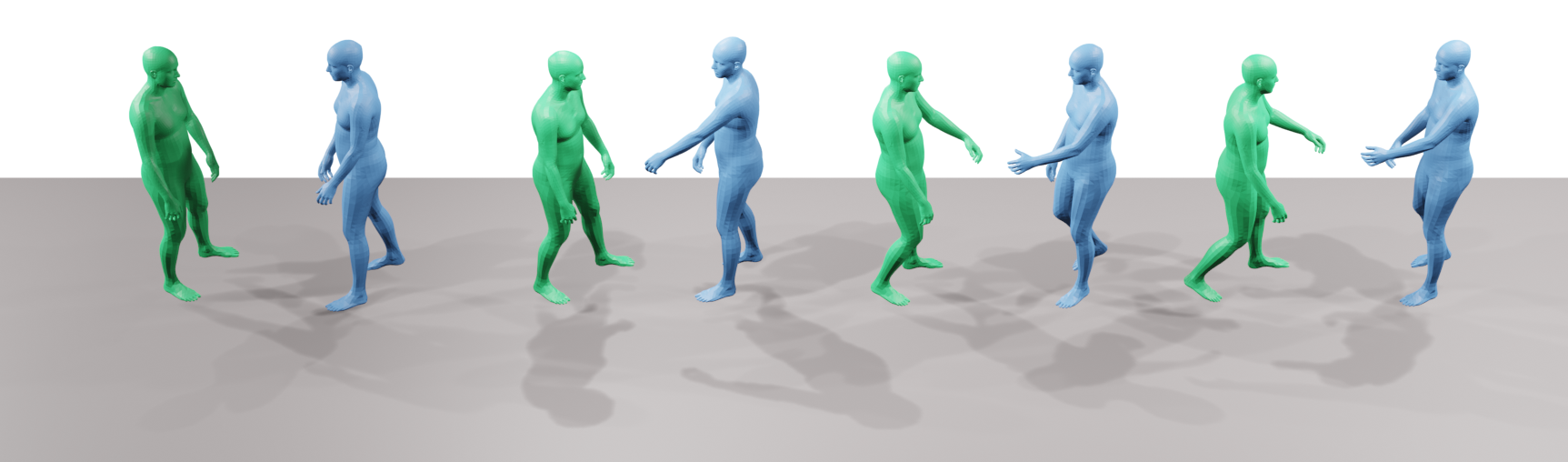}
  \caption{Failure case of our method.}
  \label{fig:fail}
\end{figure}

\noindent \textbf{Failure case.} We also show the failure cases of our motion generation pipeline in Fig.~\ref{fig:fail}. Our model cannot guarantee absolute physical authenticity, for example, ensuring that the hand contacts but does not penetrate during handshaking. Incorporating more sophisticated physical constraints may solve the failure cases and be considered in future.



\section{Broader Impacts}
\label{appendix:impact}
Our model demonstrates significant potential for AR/VR and gaming applications by enabling the generation of plausible human reactions. Beyond virtual environments, the proposed approach provides an innovative technical pathway for real-world human-robot interaction, where motion patterns can be transferred to robotic systems through motion remapping technology. Although this advancement may inspire future research, we acknowledge potential misuse risks similar to other generative models, warranting ethical considerations as the technology develops.
\end{document}